\documentclass{article}

\usepackage{iclr2027/iclr2027_conference,times}
\iclrfinalcopy

\usepackage[utf8]{inputenc}
\usepackage[T1]{fontenc}
\usepackage[colorlinks=true,citecolor=blue,linkcolor=blue,urlcolor=black,hypertexnames=false]{hyperref}
\usepackage{url}
\expandafter\def\expandafter\UrlBreaks\expandafter{\UrlBreaks\do\-}
\usepackage{booktabs}
\usepackage{amsfonts}
\usepackage{amsmath}
\usepackage{amssymb}
\usepackage{nicefrac}
\usepackage{microtype}
\usepackage{refcount}
\usepackage{graphicx}
\usepackage{xcolor}
\usepackage{algorithm}
\usepackage{algpseudocode}
\usepackage{listings}
\usepackage[most]{tcolorbox}
\lstdefinestyle{promptstyle}{
  escapechar=~,
  basicstyle=\ttfamily\footnotesize,
  breaklines=true,
  breakatwhitespace=true,
  breakindent=1.5em,
  columns=flexible,
  keepspaces=true,
  backgroundcolor=\color{gray!7},
  frame=leftline,
  framerule=1.5pt,
  rulecolor=\color{gray!55},
  framexleftmargin=4pt,
  xleftmargin=10pt,
  xrightmargin=6pt,
  aboveskip=8pt,
  belowskip=8pt,
}
\newtcblisting{promptbox}[1]{%
  listing only, breakable, enhanced,
  listing options={style=promptstyle, frame=none, backgroundcolor={}, xleftmargin=0pt, xrightmargin=0pt, aboveskip=0pt, belowskip=0pt},
  colback=blue!5, colframe=blue!40!gray, boxrule=0.6pt, arc=2pt,
  title={#1}, fonttitle=\small\bfseries, coltitle=black, colbacktitle=blue!12,
  left=4pt, right=4pt, top=3pt, bottom=3pt, before skip=8pt, after skip=8pt,
}
\lstdefinestyle{codestyle}{
  escapechar=@,
  basicstyle=\ttfamily\small,
  columns=fixed,
  keepspaces=true,
}
\newtcblisting{codebox}[1]{%
  listing only, enhanced,
  listing options={style=codestyle, frame=none, backgroundcolor={}, xleftmargin=0pt, xrightmargin=0pt, aboveskip=0pt, belowskip=0pt},
  colback=blue!5, colframe=blue!40!gray, boxrule=0.6pt, arc=2pt,
  title={#1}, fonttitle=\small\bfseries, coltitle=black, colbacktitle=blue!12,
  left=4pt, right=4pt, top=3pt, bottom=3pt, before skip=8pt, after skip=8pt,
}
\usepackage{longtable}
\usepackage{multirow}
\usepackage{tikz}
\usetikzlibrary{shapes.geometric, arrows.meta, positioning, fit, backgrounds}
\usepackage{pgfplots}
\pgfplotsset{compat=1.17}
\definecolor{cPO}{HTML}{0072B2}
\definecolor{cKG}{HTML}{E69F00}
\definecolor{cML}{HTML}{009E73}
\definecolor{cFS}{HTML}{D55E00}
\definecolor{cBEST}{HTML}{E69F00}
\definecolor{cORA}{HTML}{D55E00}

\newcommand{\MathKG}{\textsc{MathKG}}
\newcommand{\MathAgent}{\textsc{MathAgent}}
\newcommand{\Mathlib}{\textsc{Mathlib}}

\title{When Does Structured Knowledge Help \\ Neural Theorem Proving?}

\author{%
  Sareh Nabi, Roland Vogl, Marzieh Nabi\thanks{Corresponding author:
  mnabi@stanford.edu. Project page: \href{https://sarehnabi.github.io/mathagent/}{\textcolor{blue}{\nolinkurl{https://sarehnabi.github.io/mathagent/}}}.} \\
  {\normalfont\small Stanford University} \\
}

\begin{document}

\maketitle
\ificlrfinal\lhead{}\renewcommand{\headrulewidth}{0pt}\fi

\begin{abstract}
Does structured mathematical knowledge help LLMs prove theorems in Lean~4? If so, for
which models, and does the answer vary by problem? Formal libraries such as \Mathlib{} encode 285{,}000+
verified theorems with syntactic dependencies, but the semantic layer mathematicians
rely on for discovery, such as analogies, generalizations, and cross-domain bridges,
remains implicit. We introduce
\MathAgent{}, a system that builds this missing layer as a knowledge graph called
\MathKG{}, and uses it to augment LLM theorem provers. \MathKG{} connects 364 \Mathlib{}
theorems and definitions by 9{,}434 typed semantic edges, inferred via LLM-based relation
extraction anchored to verified \Mathlib{} declarations. We run a controlled ablation across four
augmentation modes (no external context, knowledge-graph context, \Mathlib{} library
retrieval, and both combined) and five models: the general-purpose Qwen3-8B/32B, their
Lean-specialized derivatives Goedel-Prover-V2-8B/32B, and the
proprietary Claude~Sonnet~4.6. We evaluate on miniF2F, adding PutnamBench and
MathOlympiadBench for the proprietary model.
Three findings emerge. (i)~\emph{Specialization dominates augmentation} by an order of
magnitude: Lean fine-tuning adds 33--38 percentage points of solve rate in \emph{every}
augmentation mode, and a specialized 8B model even beats a $4\times$ larger general one
by 29--35 points in every mode; by contrast, no single augmentation mode improves solve rate by more than
3 points, so external knowledge does not substitute for competence in the weights. (ii)~Augmentation's aggregate effect is
\emph{capability-conditioned}: knowledge-graph context helps small models but hurts large
ones, with the specialized model gaining more relative to its general base at every scale.
(iii)~Yet across every model the augmentation modes solve
\emph{different} problems, so an oracle that selects the best augmentation mode per problem solves
6\% to 58\% more problems than the unaugmented prover (least for the proprietary model,
most for a weak general one), revealing a \emph{complementarity effect} that strengthens
on harder problems (32\% more on PutnamBench). Together these results motivate
adaptive strategies that select augmentation by model capability and problem.
\ificlrfinal
Code, data, and artifacts are available at \href{https://github.com/sarehnabi/mathagent}{\textcolor{blue}{\nolinkurl{https://github.com/sarehnabi/mathagent}}}.
\else
Code, data, and artifacts will be released upon publication.
\fi
\end{abstract}

\section{Introduction}
\label{sec:intro}

Large language models are rapidly becoming capable theorem provers. Combined with
search and formal verification, LLMs now produce
IMO-level proofs~\citep{achim2025aristotle,hubert2026olympiad},
and LLM-based systems increasingly contribute to resolving open
problems~\citep{romera2024funsearch,novikov2025alphaevolve,georgiev2025exploration,tsoukalas2026advancing,openai2026tenadvances,alon2026remarks,ono2026arctan}.
Specialized open provers are built on the Lean~4 proof
assistant~\citep{moura2021lean4} and its \Mathlib{} library~\citep{mathlib2020} of
285{,}000+ theorems,\footnote{\Mathlib{} spans algebra, analysis,
combinatorics, number theory, topology, and beyond; it contains 285{,}000+ theorems and
135{,}000+ definitions~\citep{mathlib_stats2026}.} and are fine-tuned on large corpora of
machine-verified Lean proofs~\citep{ren2025deepseekproverv2,lin2026goedel}: everything they
know about mathematics is stored in their weights.

Mathematicians work differently. Beyond
what any individual knows, they draw on a shared semantic web of the field:
which theorems are analogous across subfields, which results generalize others, which
identities bridge number theory to combinatorics. As proofs become abundant,
\citet{tao2026mathematics} argues, emphasis should move from proof generation to proof
digestion: exposition, refereeing, and connecting results into a shared body of knowledge.
This paper asks: if we make this layer explicit and hand it to an LLM
prover at inference time, does it prove more theorems? If so, for which models, and does
the answer vary by problem?

Formal libraries such as \Mathlib{} encode \emph{syntactic}
structure: which theorems are used to prove others and which definitions they invoke.
The \emph{semantic} relationships that guide discovery live in textbooks, seminars, and
expert intuition. We build this layer explicitly from a curated core.
\MathKG{} is a knowledge graph (KG) of 364 \Mathlib{}-verified theorems and definitions
connected by 9{,}434 typed semantic edges in eight categories (e.g., generalizes,
analogous-to, cross-domain-bridge), inferred by LLM-based relation
extraction over the nodes' statements and verified \Mathlib{} declaration names.
\MathAgent{}, four LLM agents coordinated by an Orchestrator over \MathKG{} as shared memory,
injects knowledge-graph context, \Mathlib{} library retrieval, or both into an LLM prover with
error-guided refinement; the ablation exercises only its Explorer and Prover.

Retrieval- and KG-augmented provers report gains for particular
models (\S\ref{sec:related}), but to our knowledge no controlled study isolates the
effect of structured knowledge across matched general and Lean-specialized model pairs at
multiple scales. We therefore
evaluate every combination of four augmentation modes (none, knowledge-graph context,
lexical \Mathlib{} library retrieval, both) and five models. Four form a matched $2\times2$ of
general-purpose versus Lean-specialized provers at 8B and 32B:
Qwen3-8B/32B~\citep{qwen3_2025} versus Goedel-Prover-V2-8B/32B, each fine-tuned from its
Qwen3 base~\citep{lin2026goedel}. The fifth, Claude Sonnet~4.6~\citep{anthropic2026sonnet46},
serves as a proprietary reference. All five are evaluated at pass@32 on miniF2F, and the
proprietary model additionally on PutnamBench and MathOlympiadBench.

\paragraph{Contributions.} We make four contributions:
\begin{enumerate}
  \item \textbf{\MathKG{}}: A semantic knowledge graph of 364
        \Mathlib{}-verified theorems and definitions across seven domains with 9{,}434 typed edges in eight
        relational categories, built by an LLM relation-extraction pipeline anchored to
        verified \Mathlib{} declarations (\S\ref{sec:mathkg}).

  \item \textbf{\MathAgent{}}: Four LLM agents (Explorer, Conjecturer, Prover, Integrator)
        coordinated by an Orchestrator, for targeted proving (evaluated here) and open-ended
        exploration, with
        \MathKG{} as persistent shared memory and error-guided Lean~4 verification
        (\S\ref{sec:mathagent}).

  \item \textbf{Controlled cross-model ablation}: A systematic evaluation of four
        augmentation modes (KG context, \Mathlib{} retrieval, both, or neither) across
        matched general-purpose Qwen3 vs.\ Lean-specialized Goedel-Prover-V2 pairs at 8B and
        32B, plus the proprietary Claude
        Sonnet~4.6, on miniF2F, extended to PutnamBench and MathOlympiadBench
        (\S\ref{sec:experiments}).

  \item \textbf{Three findings on when structured knowledge helps}:
        (i)~\emph{specialization dominates augmentation}: Lean fine-tuning adds 33--38
        percentage points in every mode, a specialized 8B model beats a general model four
        times its size, and no augmentation mode improves any model's aggregate by more than
        3 points; (ii)~augmentation is \emph{capability-conditioned}: knowledge-graph context helps
        small models but hurts large ones, with the specialized model gaining more relative to
        its general base at every scale; (iii)~the aggregates hide \emph{complementarity}: the modes solve different
        problems, so an oracle selecting the best mode per problem solves 6--58\% more than
        the unaugmented prover, a margin that grows on harder benchmarks and motivates
        adaptive selection (\S\ref{sec:exp_specialization}--\ref{sec:analysis}).
\end{enumerate}
\section{Related Work}
\label{sec:related}

\paragraph{AI for theorem proving.}
Since GPT-f brought language models to formal proving~\citep{polu2020gptf}, the field has
produced headline systems such as Aristotle~\citep{achim2025aristotle} and
AlphaProof~\citep{hubert2026olympiad}. Lean-specialized provers such as
Goedel-Prover~\citep{lin2025goedelprover,lin2026goedel},
DeepSeek-Prover-V2~\citep{ren2025deepseekproverv2}, Kimina-Prover~\citep{sun2025kimina},
BFS-Prover-V2~\citep{liu2025bfsproverv2}, and Pythagoras-Prover~\citep{leang2026pythagoras}
push miniF2F upward through fine-tuning, expert iteration, RL, and tree
search~\citep{lample2022htps}. Self-play~\citep{dong2025stp,bailey2026sgs} and RL against
Lean feedback~\citep{chen2025seedprover15,chen2025seedprover} extend training beyond fixed
problem sets, and
Leanstral~1.5~\citep{kabra2026leanstral} saturates miniF2F as a general code agent.

Minerva~\citep{lewkowycz2022minerva}, Llemma~\citep{azerbayev2023llemma}, and
WizardMath~\citep{luo2025wizardmath} are trained on mathematical text and solve problems
mainly in natural language.
Draft, Sketch, and Prove~\citep{jiang2022draftsketchprove} guides
formal provers with informal sketches, autoformalization has grown from competition
statements~\citep{wu2022autoformalization} to verified libraries at
scale~\citep{rammal2026formalizing}, and
Pseudo-Formalization~\citep{barkallah2026pseudoformalization} verifies natural-language
proofs in pseudo-formal blocks.
Agentic frameworks such as Hilbert~\citep{chen2025hilbert},
HERMES~\citep{ospanov2026hermes}, and LEAP~\citep{kung2026leap} interleave informal
reasoning with iterative Lean verification, building on COPRA~\citep{thakur2024copra} and
Lemur~\citep{wu2024lemur}; MATEK~\citep{kerger2026matek} adds a persistent typed graph of
claims, proofs, and audits as cross-session memory for research agents; and general
multi-agent frameworks~\citep{wu2023autogen,hong2023metagpt} supply generic
orchestration, not mathematical reasoning.
All draw their strength from training or from agentic scaffolding; none isolates the
contribution of structured knowledge supplied at inference time, our central question.

\paragraph{Retrieval-augmented theorem proving.}
Premise selection~\citep{irving2016deepmath,wang2017premise,mikula2024magnushammer}
ranks lemmas by learned embeddings; LeanDojo~\citep{yang2023leandojo} trains the
retrieval-augmented ReProver on Lean premises, and
Lean Copilot~\citep{song2025leancopilot} integrates such models natively into Lean.
REAL-Prover~\citep{shen2025realprover} adds \Mathlib{} retrieval to a stepwise
prover, LeanSearch~v2~\citep{gao2026leansearch} retrieves whole premise sets, and
DRIFT~\citep{li2025drift} decomposes statements before retrieval, nearly doubling F1 on
ProofNet~\citep{azerbayev2023proofnet}.
Our Explorer also builds problem-level context, combining \Mathlib{} retrieval with
\MathKG{}'s typed relations and their reasoning.
\citet{khalov2026explainable} and \citet{ataeva2026multilevel} retrieve lemma hints for
DeepSeek-Prover-V2-7B over an ontology-typed dependency graph of \Mathlib{}, with gains
concentrated on problems the bare model rarely solves.
KG-Prover~\citep{wang2025kgprover} augments provers with a text-mined knowledge graph,
gaining most on smaller models; our graph is anchored to verified \Mathlib{} declarations
under a typed ontology, and our controlled ablation isolates KG, retrieval, and both,
showing when each helps or hurts and that the modes are complementary.

\paragraph{Mathematical knowledge graphs.}
NaturalProofs~\citep{welleck2021naturalproofs} links ProofWiki theorems through their
reference structure, without formal verification.
TheoremGraph~\citep{kurgan2026theoremgraph} unifies arXiv statements with Lean~4
declarations by embedding similarity, and \citet{li2026network} and
\citet{dasgupta2026geometric} map \Mathlib{}'s dependency network; all three graphs encode
dependency or citation links rather than semantic relations.
AutoMathKG~\citep{zhang2025automathkg} builds an updatable KG from
ProofWiki, textbooks, arXiv, and TheoremQA with LLM augmentation, aimed at informal reasoning.
We apply LLM-based relation
extraction~\citep{ye2022generative,wadhwa2023revisiting,xu2024generativeie} over
\Mathlib{} declarations. \MathKG{} differs in two ways: its nodes
are verified \Mathlib{} declarations, and its edges are LLM-inferred semantic relations
under an eight-type ontology.

\section{\MathKG{}: A Semantic Knowledge Graph}
\label{sec:mathkg}

\MathKG{} organizes a curated core of \Mathlib{}, 364 nodes in seven domains, into a semantic graph.

\paragraph{Node selection.} The seven domains cover the subject areas of the three
evaluation benchmarks (\S\ref{sec:setup}). Within each domain we asked Claude Opus~4.7,
Gemini~3.1~Pro, and GPT-5.4 for the fundamental theorems and definitions of a standard
undergraduate mathematics curriculum, cross-checked the lists against one another, and kept
only entries whose formalization exists in \Mathlib{}. The lists were also reviewed by a PhD
mathematician, who judged each a reasonable set of fundamentals. This yields 229 theorems and
135 definitions, 49 to 55 nodes per domain (theorems/definitions): number theory 39/10,
combinatorics 33/20, algebra 34/18, abstract algebra 30/25, linear algebra 28/24, analysis
40/14, and Euclidean geometry 25/24. The core is
deliberately small: a graph over all of \Mathlib{} would have over $10^{11}$
ordered pairs, beyond exhaustive relation extraction, and would bury the fundamentals
under specialized lemmas. A fundamentals-only graph keeps exhaustive pairwise
extraction feasible (132{,}132 calls) and leaves specialized lemmas to \Mathlib{}
retrieval (\S\ref{sec:mathagent}).

\subsection{Graph Schema}

\MathKG{} is a directed graph $\mathcal{G} = (\mathcal{V}, \mathcal{E})$, built with
NetworkX,\footnote{NetworkX is a standard Python library for in-memory graphs (details in
Appendices~\ref{app:edge_cleanup} and~\ref{app:implementation}).} where:

\paragraph{Nodes.} Each node is a mathematical object with an identifier, a
human-readable name, a domain, a natural-language statement, a type (theorem or
definition), keywords, its \Mathlib{} declaration name and import, and a verification
status. Every declaration name is verified against the 404{,}440-entry \Mathlib{}
declaration database, so each node names a theorem or definition formalized in
\Mathlib{}.\footnote{Appendix~\ref{app:mathkg_node} shows example nodes; Appendix~\ref{app:mathlib_primer}
introduces Lean~4, \Mathlib{}, and the declaration database.}

\paragraph{Edges.} Each directed edge from node $A$ to node $B$, encoded as
$(A, B, r, c, \rho)$, carries a relation type $r \in \mathcal{R}$, a confidence
$c \in [0,1]$, and a natural-language justification $\rho$. The eight edge types
(Table~\ref{tab:edge_types}) carry node-type restrictions (Appendix~\ref{app:edge_restrictions})
and range from logical relations (generalizes, specializes, equivalent-to) to
discovery-oriented ones (analogous-to, cross-domain-bridge).

\begin{table}[h]
\centering
\caption{The eight semantic edge types (directed edge $A \to B$) in \MathKG{}.}
\label{tab:edge_types}
\small
\setlength{\tabcolsep}{1.5pt}
\resizebox{\textwidth}{!}{%
\begin{tabular}{llll}
\toprule
\textbf{Edge Type} & \textbf{Semantics} & \textbf{$A$/$B$ Restriction} & \textbf{Example} \\
\midrule
prerequisite        & $A$ is needed to state $B$         & $A$: definition     & Prime Number $\to$ Fermat's LT \\
generalizes         & $A$ is more general than $B$       & both theorems       & Binom.\ Thm $\to$ Sum of Binom.\ Coeff. \\
specializes         & $A$ is a special case of $B$       & both theorems       & Fermat's LT $\to$ Lagrange's Thm \\
equivalent-to       & $A \leftrightarrow B$, same level  & both theorems       & Open Mapping $\leftrightarrow$ Closed Graph \\
applied-in          & $A$'s result used in $B$'s proof   & both theorems       & FTA $\to$ Lifting the Exp.\ Lemma \\
analogous-to        & Same structure, other context      & both theorems       & Hensel $\leftrightarrow$ Impl.\ Fn.\ Thm \\
cross-domain-bridge & Formal link across domains         & different domains   & Vector Space $\to$ Field \\
instance-of         & Concrete $A$ exemplifies $B$       & both definitions    & ZMod $n$ $\to$ Ring \\
\bottomrule
\end{tabular}}
\end{table}

\subsection{Relation Extraction Pipeline}
\label{sec:relation_extraction}

Given an ordered pair of nodes $(A, B)$, we prompt Claude Sonnet~4.6 to infer the directed
semantic relationship $(A \to B)$. The prompt provides both nodes' natural-language
names and statements, keywords and domain labels, and \Mathlib{} declaration names and
imports, together with definitions, usage restrictions, and an illustrative pair for each
of the eight edge types, plus few-shot examples (the complete prompt is in
Appendix~\ref{app:relation_extraction}). Hard type restrictions (e.g., generalizes requires
both endpoints to be theorems) are stated in the prompt and enforced again in
post-processing, and a ``none'' fallback lets the model reject weak connections.

The model returns a JSON object $\{\text{relationship}: r,\ \text{confidence}: c,\
\text{reasoning}: \rho\}$; edges with $c < \tau$ (default $\tau = 0.75$) are discarded.
We check both directions
$(A \to B)$ and $(B \to A)$ for all pairs, as relationships are generally asymmetric.
A post-processing pass then (i) removes edges that violate the type restrictions;
(ii) removes contradictory bidirectional generalizes/specializes cycles; and (iii) fills in
inverses for the symmetric types and the generalizes/specializes pair.

\section{\MathAgent{}: Four-Agent Discovery System}
\label{sec:mathagent}

\begin{figure}[h]
\centering
\scalebox{0.92}{%
\begin{tikzpicture}[
  node distance=1.8cm,
  agent/.style={rectangle, rounded corners, draw=black, fill=blue!10,
                text width=2.8cm, align=center, minimum height=1.2cm, font=\normalsize\bfseries},
  action/.style={rectangle, draw=gray!60, fill=gray!10,
                 text width=2.6cm, align=center, minimum height=0.8cm, font=\tiny},
  arrow/.style={-{Stealth[length=6pt]}, thick},
  graph/.style={cylinder, shape border rotate=90, draw=orange!80, fill=orange!15,
                minimum height=2.2cm, minimum width=1.2cm, aspect=0.3, align=center, font=\small\bfseries},
]
\node[graph] (kg) at (0,0) {{\normalsize\MathKG}\\[3pt](Shared Memory)};
\node[agent] (explorer)    at (-4,  2) {Explorer};
\node[agent] (conjecturer) at (-4, -2) {Conjecturer};
\node[agent] (prover)      at ( 4, -2) {Prover};
\node[agent] (integrator)  at ( 4,  2) {Integrator};
\draw[-{Stealth[length=8pt,width=6pt]}, line width=1.1pt, color=blue!70] (explorer)    -- node[left,  font=\normalsize] {targets}       (conjecturer);
\draw[-{Stealth[length=8pt,width=6pt]}, line width=1.1pt, color=blue!70] (conjecturer) -- node[below, font=\normalsize] {conjecture}    (prover);
\draw[-{Stealth[length=8pt,width=6pt]}, line width=1.1pt, color=blue!70] (prover)      -- node[right, font=\normalsize] {proof result}  (integrator);
\draw[-{Stealth[length=8pt,width=6pt]}, line width=1.1pt, color=blue!70] (integrator)  -- node[above, font=\normalsize] {enriched graph} (explorer);
\draw[-{Stealth[length=9pt,width=7pt]}, thick, color=orange!80, dashed] (kg) -- (explorer);
\draw[-{Stealth[length=9pt,width=7pt]}, thick, color=orange!80, dashed] (kg) -- (conjecturer);
\draw[-{Stealth[length=9pt,width=7pt]}, thick, color=orange!80, dashed] (kg) -- (prover);
\draw[{Stealth[length=9pt,width=7pt]}-{Stealth[length=9pt,width=7pt]}, thick, color=orange!80, dashed] (kg) -- (integrator);
\end{tikzpicture}%
}
\caption{The \MathAgent{} architecture: four LLM agents with \MathKG{} as persistent shared
memory (the Orchestrator, which coordinates the cycle, is omitted).}
\label{fig:architecture}
\end{figure}
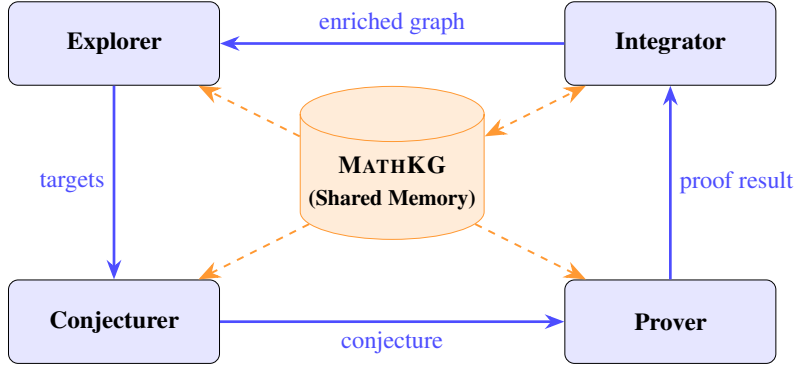

\MathAgent{} implements a four-agent architecture (Figure~\ref{fig:architecture}) where
\MathKG{} serves as the shared persistent memory substrate. The four agents,
\emph{Explorer}, \emph{Conjecturer}, \emph{Prover}, and \emph{Integrator}, are
LLM-powered modules with specialized roles that share access to the graph.

The system supports two operation modes. In \emph{open-ended exploration}, the Explorer
identifies investigation targets (structural gaps, cross-domain analogies) in \MathKG{},
the Conjecturer proposes a candidate theorem, the Prover attempts a Lean~4 proof, and the
Integrator evaluates novelty and writes verified results back into the graph. This mode is
implemented but not evaluated in this paper.

\emph{Targeted problem-solving}, which is the focus of our work, is shown in
Algorithm~\ref{alg:discovery}: the Explorer searches
\MathKG{} and \Mathlib{} for context relevant to a given problem and the Prover attempts a
proof using that context. In both modes a deterministic \emph{Orchestrator} sequences the
agents; in targeted mode it also feeds failed approaches back as negative context.

\begin{algorithm}[h]
\caption{\MathAgent{} Targeted Problem-Solving Mode}
\label{alg:discovery}
\begin{algorithmic}[1]
\Require \MathKG{} $\mathcal{G}$, problem $P$, max attempts $K$ (default 32)
\State $\text{context}_{\text{KG}} \gets \text{Explorer.search\_kg}(P, \mathcal{G})$ \Comment{Layer 1: KG semantic search}
\State $\text{context}_{\text{Mathlib}} \gets \text{Explorer.search\_mathlib}(P)$ \Comment{Layer 2: Mathlib declaration search}
\State $\text{proof} \gets \text{Prover.generate}(P, \text{context}_{\text{KG}}, \text{context}_{\text{Mathlib}})$ \Comment{Attempt 1: context injected once}
\For{$k = 2, \ldots, K$} \Comment{Error-guided refinement}
  \State \textbf{if} $\text{proof.verified}$ \textbf{then} \Return $\text{proof}$
  \State $\text{proof} \gets \text{Prover.refine}(\text{proof}, \text{proof.error\_log})$ \Comment{Previous code + its Lean errors}
\EndFor
\State \Return $\text{proof}$ if $\text{proof.verified}$ else $\text{failed}$
\end{algorithmic}
\end{algorithm}

\noindent The four augmentation conditions evaluated in \S\ref{sec:experiments}
correspond to lines of Algorithm~\ref{alg:discovery}: \texttt{prover\_only} skips lines 1--2 (no external context);
\texttt{with\_kg} executes only line 1 (\MathKG{} semantic context);
\texttt{with\_mathlib} executes only line 2 (\Mathlib{} library retrieval context);
\texttt{full\_system} executes both (\MathKG{} + \Mathlib{} context combined).

\subsection{Explorer Agent}
\label{sec:agents}

In targeted problem-solving, the Explorer performs a \emph{two-layer search} to gather the
Prover's context.

\paragraph{Layer 1: \MathKG{} semantic search.} The Explorer (1) pre-filters \MathKG{} to
nodes in the problem's domain plus their cross-domain-bridge neighbors, (2) selects
$s{=}5$ seeds by LLM relevance scoring, (3) walks \MathKG{} edges one hop from each seed,
and (4) attaches to each retrieved node the edges that connect it to the seeds. The
output is a context block of the seeds and their one-hop neighbors, each with its
statement and a relevance rationale, and each neighbor with its connecting edges (type,
confidence, and reasoning); details are in Appendix~\ref{app:explorer}.

\paragraph{Layer 2: \Mathlib{} declaration search.} The Explorer (1) extracts search terms
(Lean~4 identifiers regex-extracted from the formal statement plus LLM-suggested
\Mathlib{} namespace prefixes), (2) prefix-matches them against the declaration names of the
\Mathlib{} documentation index (\S\ref{sec:mathkg}), (3) scores and ranks the matches by source priority, match precision, name
specificity, declaration kind, and module proximity, and (4) enriches the top-ranked
declarations with their Lean~4 type signatures and docstrings, so the Prover receives
exact names and argument types (details in Appendix~\ref{app:explorer}).

\paragraph{Full system (Layers 1+2).} When both layers run, their context blocks are
concatenated, knowledge-graph context first; Layer~2 then favors modules of the KG seeds and
drops declarations the KG block already lists (Appendix~\ref{app:explorer}).

\subsection{Prover Agent}
\label{sec:prover}

The Prover attempts a complete Lean~4 proof with error-guided refinement:

\begin{enumerate}
  \item \textbf{Initial generation}: Prompt the prover LLM (the evaluated model,
        \S\ref{sec:setup}) with the natural-language and Lean~4 statements plus the
        Explorer's context (targeted problem-solving) or the Conjecturer's proof sketch
        (open-ended exploration); prompts are in Appendix~\ref{app:prover_prompts}.
  \item \textbf{Lean~4 verification}: Submit the generated code to the Lean~4 type-checker.
  \item \textbf{Error-guided refinement}: On failure, feed the previous code and its error
        messages (up to ten) back to the model to fix. Repeat steps 2--3 until success or $K$
        attempts.\footnote{We use $K{=}32$ for benchmark evaluation (pass@32) in \S\ref{sec:experiments}.}
\end{enumerate}

\section{Experiments}
\label{sec:experiments}

We run a controlled ablation to determine when structured knowledge augmentation helps an
LLM theorem prover, for which models, and by how much: five models under four
augmentation modes, isolating the contributions of the \MathKG{} semantic graph,
\Mathlib{} library retrieval, and their combination. \S\ref{sec:exp_specialization} asks
whether augmentation or Lean specialization matters more;
\S\ref{sec:exp_conditioned} whether augmentation's effect depends on the model that
receives it; and \S\ref{sec:analysis} whether different modes solve different problems,
so that choosing the augmentation mode per problem would beat any fixed mode.

\subsection{Setup}
\label{sec:setup}
\label{sec:benchmark}

\paragraph{\MathKG{}.}
\MathKG{} (\S\ref{sec:mathkg}) contributes 364 nodes and 9{,}434 edges: 8{,}026
LLM-inferred plus 1{,}408 auto-inverse. Of the 8{,}026 LLM-inferred edges,
cross-domain-bridge (3{,}481) and prerequisite (2{,}293) account for 72\%, applied-in and
analogous-to for most of the rest (1{,}503 and 578), and the logical types are infrequent
(instance-of 80, generalizes 48, specializes 30, equivalent-to 13); connectivity statistics
are in Appendix~\ref{app:edge_cleanup}.

\paragraph{Benchmarks.}
We evaluate on three standard formal-mathematics benchmarks:
\textbf{miniF2F}~\citep{zheng2022minif2f} (488 olympiad-level problems: the full
validation and test splits),
\textbf{MathOlympiadBench}~\citep{lin2026goedel} (360 competition problems),
and \textbf{PutnamBench}~\citep{tsoukalas2024putnambench} (672 undergraduate/%
research-level problems). The cross-model comparison uses miniF2F;
\S\ref{sec:complementarity} adds the harder benchmarks for the proprietary model.

\paragraph{Augmentation modes.}
Four conditions isolate each component (Algorithm~\ref{alg:discovery}):
\texttt{prover\_only} (no external context),
\texttt{with\_kg} (\MathKG{} semantic search, Layer~1),
\texttt{with\_mathlib} (\Mathlib{} library retrieval, Layer~2), and
\texttt{full\_system} (both). Comparing each augmented mode against the baseline
isolates each component's contribution; the combined mode tests for synergy.

\paragraph{Models.}
We compare five models along two axes, general-purpose vs.\ Lean-specialized and
small (8B) vs.\ large (32B), forming a controlled $2\times2$: Qwen3-8B and Qwen3-32B
(general)~\citep{qwen3_2025} versus Goedel-Prover-V2-8B and -32B
(Lean-specialized)~\citep{lin2026goedel}.
Goedel-Prover-V2-8B is Qwen3-8B fine-tuned on Lean proofs through expert iteration and
self-correction training, and -32B is fine-tuned from Qwen3-32B, so each pair is matched in
scale and architecture. We add Claude~Sonnet~4.6~\citep{anthropic2026sonnet46} as a
proprietary, larger reference.

\paragraph{Protocol.}
We report pass@32 with sequential, error-guided refinement (Algorithm~\ref{alg:discovery})
rather than 32 independent samples.
Sonnet~4.6 runs on AWS~Bedrock and the four open-weight models on local vLLM;
only the model and serving backend change, with benchmark sets and augmentation modes
held fixed. Every LLM call in a run, including the Explorer's relevance scoring and
search-term suggestion, uses that run's model, so each result measures the end-to-end
system a practitioner would deploy with that model. All five models use the same generic prompt (Appendix~\ref{app:prover_prompts}); only the
injected context varies across modes. We use temperature 0.3 for initial
generation and 0.2 for refinement, and set each model's maximum output tokens above the
length of its longest verified proof (per-model caps in Appendix~\ref{app:implementation}). Lean~4 compilation has a 1{,}200\,s
timeout per attempt. The Explorer uses the same default settings for every model
(\S\ref{sec:agents}, Appendix~\ref{app:explorer}).

\subsection{Specialization dominates knowledge augmentation}
\label{sec:exp_specialization}

\begin{figure}[!h]
\vspace{-4pt}
\centering
\begin{tikzpicture}
\begin{axis}[
  ybar,
  bar width=10pt,
  width=\textwidth,
  height=6.3cm,
  enlarge x limits=0.12,
  ymin=0, ymax=95,
  ylabel={Solve Rate (\%)},
  ylabel style={font=\small},
  symbolic x coords={Qwen3-8B, Qwen3-32B, Goedel-8B, Goedel-32B, Sonnet 4.6},
  xtick=data,
  xtick pos=bottom,
  x tick label style={font=\small},
  ytick={0,20,40,60,80},
  tick label style={font=\small},
  ymajorgrids=true,
  grid style={gray!20, densely dotted, line width=0.3pt},
  area legend,
  legend cell align=left,
  nodes near coords,
  every node near coord/.append style={font=\footnotesize, rotate=90, anchor=west,
    /pgf/number format/.cd, fixed, fixed zerofill, precision=1},
  legend style={at={(0.02,0.98)}, anchor=north west, legend columns=1, font=\small,
    draw=none, fill=none},
]
\addplot[fill=cPO, draw=cPO!70!black] coordinates {(Qwen3-8B,6.8) (Qwen3-32B,12.7) (Goedel-8B,42.0) (Goedel-32B,48.6) (Sonnet 4.6,74.8)};
\addplot[fill=cKG, draw=cKG!70!black] coordinates {(Qwen3-8B,7.2) (Qwen3-32B,9.8) (Goedel-8B,44.9) (Goedel-32B,48.0) (Sonnet 4.6,73.4)};
\addplot[fill=cML, draw=cML!70!black] coordinates {(Qwen3-8B,6.4) (Qwen3-32B,12.7) (Goedel-8B,41.8) (Goedel-32B,48.2) (Sonnet 4.6,73.6)};
\addplot[fill=cFS, draw=cFS!70!black] coordinates {(Qwen3-8B,7.4) (Qwen3-32B,9.4) (Goedel-8B,42.0) (Goedel-32B,42.6) (Sonnet 4.6,75.0)};
\legend{prover\_only, with\_kg, with\_mathlib, full\_system}
\end{axis}
\end{tikzpicture}
\vspace{-6pt}
\caption{Cross-model pass@32 on miniF2F (Table~\ref{tab:xmodel_main}).}
\label{fig:xmodel_bar}
\end{figure}
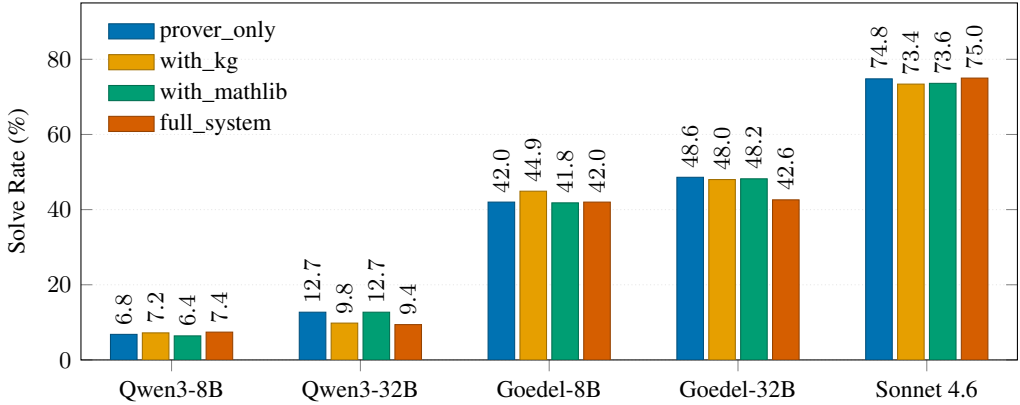

Table~\ref{tab:xmodel_main} and Figure~\ref{fig:xmodel_bar} report pass@32 on miniF2F for
all five models under the four augmentation modes. The dominant factor is neither scale nor augmentation but
\emph{specialization}: Goedel-8B solves 205/488 (42.0\%) versus 62/488 (12.7\%) for
Qwen3-32B without external context, a 29-point gap, even though Goedel-8B is four times
smaller. Within each matched pair, specialization adds 35.2 points
at 8B (42.0\% vs.\ 6.8\%) and 35.9 points at 32B (48.6\% vs.\ 12.7\%) in
\texttt{prover\_only}. By contrast, no single augmentation
mode improves solve rate by more than 3 percentage points.\footnote{Appendix~\ref{app:proofs}
shows two verified proofs of one miniF2F problem, by Sonnet~4.6 and by Goedel-8B.}

\begin{table}[h]
\centering
\caption{Cross-model pass@32 solve rates on miniF2F (488 problems). Best mode in \textbf{bold}.}
\label{tab:xmodel_main}
\vspace{3pt}
\small
\begin{tabular}{llcccc}
\toprule
\textbf{Type} & \textbf{Model} & \textbf{prover\_only} & \textbf{with\_kg} & \textbf{with\_mathlib} & \textbf{full\_system} \\
\midrule
\multirow{2}{*}{General}
 & Qwen3-8B  & 33 (6.8\%)  & 35 (7.2\%)  & 31 (6.4\%)  & \textbf{36 (7.4\%)} \\
 & Qwen3-32B & \textbf{62 (12.7\%)} & 48 (9.8\%) & \textbf{62 (12.7\%)} & 46 (9.4\%) \\
\midrule
\multirow{2}{*}{Specialized}
 & Goedel-8B  & 205 (42.0\%) & \textbf{219 (44.9\%)} & 204 (41.8\%) & 205 (42.0\%) \\
 & Goedel-32B & \textbf{237 (48.6\%)} & 234 (48.0\%) & 235 (48.2\%) & 208 (42.6\%) \\
\midrule
Proprietary & Claude Sonnet~4.6 & 365 (74.8\%) & 358 (73.4\%) & 359 (73.6\%) & \textbf{366 (75.0\%)} \\
\bottomrule
\end{tabular}
\end{table}

This dominance holds across every mode: the specialization gap (the
Lean-specialized model minus its general base at matched scale) never falls below $\sim$33
percentage points, and \texttt{with\_kg} in fact \emph{widens} it (to 37.7 points at 8B
and 38.1 at 32B). KG context gains the specialized model more relative to its general base
at both scales: $+14$ vs $+2$ at 8B, and at 32B $-3$ vs $-14$ as reported, or $+1$ vs
$-14$ once overflow is excluded.\footnote{For Goedel-32B, 10 problems in \texttt{with\_kg} and 28 in \texttt{full\_system}
had KG-augmented prompts that exceeded the served 40{,}960-token context window; we count
them as failures. Excluding them, the rates would be
49.0\% and 45.2\% (Appendix~\ref{app:overflow}). Appendix~\ref{app:mcnemar} gives McNemar $p$-values for the deltas of Table~\ref{tab:xmodel_decomp}.\label{fn:overflow}} Even against a
four-times-larger general model, a specialized 8B model
(Goedel-8B) leads Qwen3-32B by $29$--$35$ percentage points in every mode. No augmentation
mode we tested lets a general model close the gap that Lean fine-tuning opens.

\subsection{Augmentation is capability-conditioned}
\label{sec:exp_conditioned}

Whether augmentation helps \emph{on aggregate} depends on the model.
Table~\ref{tab:xmodel_decomp} decomposes each component's effect as solve-count deltas.
KG augmentation helps the specialized small model most (Goedel-8B, $+14$), helps the
weak general model marginally (Qwen3-8B, $+2$), is roughly neutral on the large
specialized model (Goedel-32B, $-3$)\hyperref[fn:overflow]{\footnotemark[\getrefnumber{fn:overflow}]}, and \emph{hurts}
the large general model (Qwen3-32B,
$-14$) and the proprietary model (Sonnet, $-7$). \Mathlib{} retrieval never helps on
aggregate ($-6$ to $0$). Combining both shows no consistent synergy: \texttt{full\_system}
improves on the best single augmentation for only two models (Qwen3-8B $+1$, Sonnet $+7$),
by 1--3 problems over \texttt{prover\_only}, and elsewhere falls below the best single mode ($-14$
to $-27$; $p$-values in Appendix~\ref{app:mcnemar}).\hyperref[fn:overflow]{\footnotemark[\getrefnumber{fn:overflow}]}

KG augmentation's effect is governed jointly by scale and specialization: scale decides the sign (KG context helps at
8B and is neutral or harmful at 32B), while specialization decides how much better than the general baseline a
model does ($+14$ vs $+2$ at 8B; $-3$ vs $-14$ at 32B). KG augmentation is thus
\emph{capability-conditioned}: its value depends on the model that receives it, so it
does not help uniformly across models. This echoes retrieval-augmented question answering, where
retrieval helps most when the model's parametric knowledge is weakest and can hurt when
the model already knows the answer~\citep{mallen2023trust}; we observe the same
conditionality for structured mathematical knowledge in formal proving.
TheoremBench~\citep{pham2026theorembench} independently finds context gains
model-dependent, and \citet{khalov2026explainable} and \citet{ataeva2026multilevel} find that graph-retrieved
hints lift DeepSeek-Prover-V2-7B only on problems it rarely solves unaided.

\Mathlib{} retrieval, by contrast, is \emph{uniformly unhelpful}, so its effect is not
conditioned on the model; this holds for our lexical retriever, which prefix-matches
declaration names (\S\ref{sec:agents}), and does not speak to learned premise
selection~\citep{yang2023leandojo,mikula2024magnushammer}. The two sources also differ in kind: knowledge-graph context is
natural-language statements and typed relations, while \Mathlib{} retrieval supplies formal
signatures. No fixed mode helps every model, nor is this a miniF2F artifact: for Sonnet it
holds across the three benchmarks, where every fixed
mode lowers the aggregate by 2.1--3.9\% (Appendix~\ref{app:sonnet_ablation}).

\begin{table}[h!]
\centering
\caption{Solve-count deltas on miniF2F (positive: augmented mode solves more); $^{*}$: $p<.05$.}
\label{tab:xmodel_decomp}
\vspace{3pt}
\small
\setlength{\tabcolsep}{5pt}
\begin{tabular}{lccccc}
\toprule
\textbf{Question} & \textbf{Qwen3-8B} & \textbf{Qwen3-32B} & \textbf{Goedel-8B} & \textbf{Goedel-32B} & \textbf{Sonnet} \\
\midrule
KG helps? (kg $-$ PO)                 & $+2$ & $-14^{*}$ & $+14^{*}$ & $-3$ & $-7$ \\
Mathlib helps? (mathlib $-$ PO)       & $-2$ & $0$   & $-1$  & $-2$  & $-6$ \\
KG beyond Mathlib? (full $-$ mathlib) & $+5$ & $-16^{*}$ & $+1$  & $-27^{*}$ & $+7$ \\
Synergy? (full $-$ max(kg,mathlib))   & $+1$ & $-16^{*}$ & $-14^{*}$ & $-27^{*}$ & $+7$ \\
\midrule
Best fixed mode & full & PO/mathlib & KG & PO & full \\
\bottomrule
\end{tabular}
\end{table}

\subsection{Complementarity: augmentation modes solve different problems}
\label{sec:analysis}
\label{sec:complementarity}

Aggregate solve rates understate augmentation's value, because the modes solve
\emph{different} problems. Table~\ref{tab:xmodel_oracle} reports, per model on miniF2F, the number of problems solved
(solve rate in parentheses) under \texttt{prover\_only} (no context), the best single mode
(the best fixed augmentation mode for that model), and the oracle (a selector that picks the
best mode per problem, equivalent in solve count to the union of the four modes). For every model the oracle exceeds \texttt{prover\_only}: $+23$ for Sonnet
(365$\to$388, 79.5\%), $+36$ for Goedel-8B (205$\to$241, 49.4\%), $+29$ for Goedel-32B
(237$\to$266, 54.5\%), and $+19$/$+18$ for the two general models.
The oracle uses four runs' worth of attempts; at a matched budget, with 8 attempts per mode
(32 in total, the same as \texttt{prover\_only} at pass@32), the union still exceeds
\texttt{prover\_only} for every model: $+5$ for Sonnet, $+27$ for Goedel-8B, $+22$ for
Goedel-32B, and $+18$/$+17$ for the two general models (matched-budget table in Appendix~\ref{app:matched_oracle}).

\vspace{-6pt}
\begin{table}[h]
\centering
\caption{Complementarity on miniF2F (488 problems). Best single mode: the best fixed
augmentation mode per model. Oracle: union of problems solved by any of the four modes.
Last column: oracle minus \texttt{prover\_only} (no context), in problems and solve-rate
points.}
\label{tab:xmodel_oracle}
\vspace{3pt}
\small
\begin{tabular}{lcccc}
\toprule
\textbf{Model} & \textbf{prover\_only} & \textbf{best single mode} & \textbf{oracle} & \textbf{oracle $-$ prover\_only} \\
\midrule
Qwen3-8B      & 33 (6.8\%)   & 36 (7.4\%)   & 52 (10.7\%)  & $+19$ ($+3.9\%$) \\
Qwen3-32B     & 62 (12.7\%)  & 62 (12.7\%)  & 80 (16.4\%)  & $+18$ ($+3.7\%$) \\
Goedel-8B  & 205 (42.0\%) & 219 (44.9\%) & 241 (49.4\%) & $+36$ ($+7.4\%$) \\
Goedel-32B & 237 (48.6\%) & 237 (48.6\%) & 266 (54.5\%) & $+29$ ($+5.9\%$) \\
Claude Sonnet~4.6 & 365 (74.8\%) & 366 (75.0\%) & 388 (79.5\%) & $+23$ ($+4.7\%$) \\
\bottomrule
\end{tabular}
\end{table}

\begin{figure}[h!]
\centering
\begin{tikzpicture}
\begin{axis}[
  ybar,
  bar width=15pt,
  width=0.92\textwidth,
  height=5.5cm,
  enlarge x limits=0.12,
  ymin=0, ymax=70,
  ylabel={Relative gain vs.\ prover\_only (\%)},
  ylabel style={font=\small},
  symbolic x coords={Qwen3-8B, Qwen3-32B, Goedel-8B, Goedel-32B, Sonnet 4.6},
  xtick=data,
  xtick pos=bottom,
  x tick label style={font=\small},
  tick label style={font=\small},
  ymajorgrids=true,
  grid style={gray!20, densely dotted, line width=0.3pt},
  nodes near coords={+\pgfmathprintnumber[precision=0]{\pgfplotspointmeta}\%},
  every node near coord/.append style={font=\small},
]
\addplot[fill=cORA, draw=cORA!70!black] coordinates {(Qwen3-8B,58) (Qwen3-32B,29) (Goedel-8B,18) (Goedel-32B,12) (Sonnet 4.6,6)};
\node[anchor=north east, font=\normalsize, text=gray!50!black] at (rel axis cs:0.96,0.86)
  {Relative gain $= \dfrac{\text{oracle} \,-\, \text{prover\_only}}{\text{prover\_only}}$};
\end{axis}
\end{tikzpicture}
\caption{Relative gain in the number of problems solved by the oracle (union of the four
modes) over prover\_only (no augmentation) on miniF2F; the gain shrinks as model
capability grows.}
\label{fig:oracle_relgain}
\end{figure}
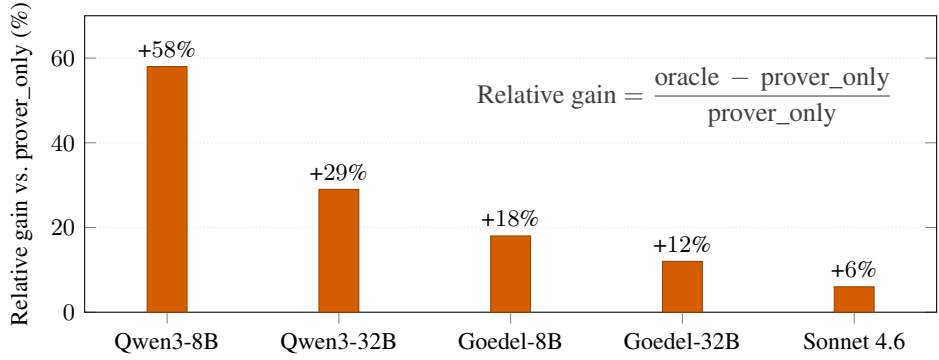

Figure~\ref{fig:oracle_relgain} plots these gains as \emph{relative} increases over
\texttt{prover\_only}: $+58\%$ for Qwen3-8B, $+29\%$ for Qwen3-32B, $+18\%$ for
Goedel-8B, $+12\%$ for Goedel-32B, and $+6\%$ for Sonnet. The oracle beats
the no-context \texttt{prover\_only} for every model, even Sonnet, where no fixed mode improves the
aggregate by more than one problem; the gain shrinks as capability grows but never vanishes. In absolute
terms (Table~\ref{tab:xmodel_oracle}), per-problem selection adds 4--7 points of solve rate.

\begin{figure}[h!]
\centering
\includegraphics[width=\textwidth]{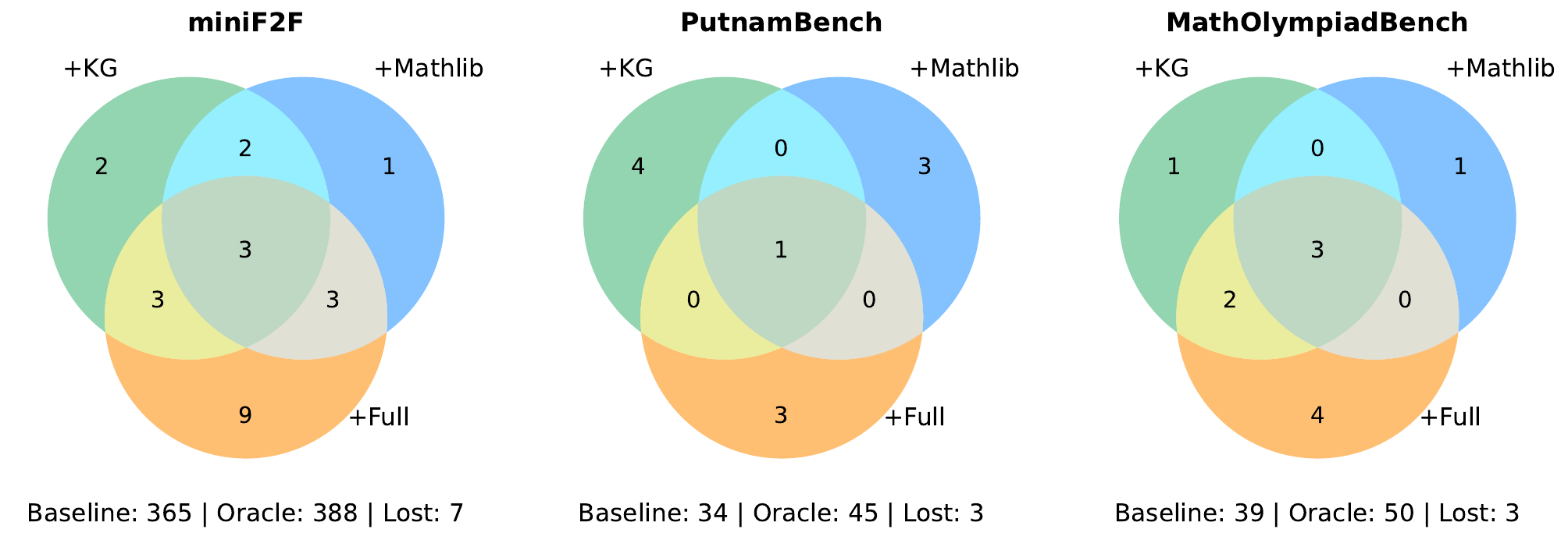}
\vspace{-8pt}
\caption{Complementarity for Sonnet~4.6: problems recovered beyond \texttt{prover\_only} by
each augmentation mode combination. Oracle: solved by any mode; Lost: solved only by
\texttt{prover\_only}.}
\label{fig:venn_main}
\end{figure}

This complementarity is not specific to miniF2F, and it \emph{strengthens} on harder
problems. For Sonnet across all three benchmarks, each augmented mode recovers a different
set of problems (Figure~\ref{fig:venn_main}): the oracle exceeds the
\texttt{prover\_only} baseline by $+6\%$ on miniF2F, $+28\%$ on MathOlympiadBench, and
$+32\%$ on PutnamBench, so an adaptive augmentation selector gains most on the two harder
benchmarks (full four-mode results in Appendix~\ref{app:sonnet_ablation}). Across the three
benchmarks, augmentation recovers 45 problems that
\texttt{prover\_only} misses and loses 13, roughly a 3:1 ratio.

\section{Conclusion}
\label{sec:conclusion}

We set out to measure whether, and when, structured mathematical knowledge helps LLMs
prove theorems in Lean~4. In a controlled ablation of four augmentation modes (no context,
knowledge-graph context, lexical \Mathlib{} retrieval, and both) across matched general and
Lean-specialized provers at 8B and 32B, plus a proprietary larger reference, three lessons
emerge. The first concerns where a prover's knowledge lives. We find that what a
model knows in its weights matters far more than what we supply at inference time: Lean
fine-tuning is worth 33 to 38 percentage points of solve rate in every mode, enough that a
specialized 8B model outperforms a general model four times its size, while no augmentation
mode improves solve rate by more than three percentage points. Structured knowledge is no
substitute for specialization.

Whether augmentation helps at all depends on the model: knowledge-graph context lifts
small provers but hurts larger ones, with the specialized model gaining more relative to its
general base at every scale. Yet this aggregate picture is incomplete. Beneath it, the
augmentation modes are genuinely complementary: they solve different problems, so a
selector that picks the best mode per problem would solve 6 to 58\% more than the
unaugmented prover, a margin that widens on harder problems (32\% more on PutnamBench). The
value of augmentation is real but latent, unlocked by selection rather than by uniform
application. This complementarity points toward adaptive proving that chooses augmentation
by model capability and problem. We release all code, graph data, and evaluation infrastructure for
the AI-for-math community.

\ificlrfinal\else
\subsection*{AI use statement}
In this work, we used generative AI tools (Claude via Claude Code) to assist in
implementing the experiment pipeline and in setting up the vLLM serving stack and GPU
instances on which the evaluated models ran. We have not used generative AI tools for
research ideation, for formulating hypotheses or theoretical frameworks, or for
formulating or proving mathematical claims; the Lean~4 proofs reported here are outputs of
the evaluated prover systems, verified by the Lean~4 kernel. Generating synthetic datasets,
translation, and qualitative data analysis are not applicable to this work; the use of LLMs
inside the method itself (node curation and relation extraction for \MathKG{}) is described in
\S\ref{sec:mathkg}. Additionally, we used generative AI tools to polish the writing. We have
reviewed all AI-assisted work: two authors checked all AI-assisted code line by line and
tested its correctness, and all text was reviewed and edited by the authors. We take
responsibility for the final content of this work, including text, claims, or artifacts
produced with the aid of generative AI.

\subsection*{Reproducibility statement}
Section~\ref{sec:experiments} and Appendix~\ref{app:implementation} specify the models,
decoding parameters, per-model token caps, Lean~4 and \Mathlib{} toolchain versions,
hardware, and the pass@32 protocol; Appendices~\ref{app:edge_prompt} and
\ref{app:prompts} reproduce the relation-extraction, Explorer, and Prover prompts verbatim;
Appendix~\ref{app:explorer} describes the retrieval procedure. We release the complete implementation, the \MathKG{}
graph file (364 nodes, 9{,}434 edges), the benchmark evaluation scripts, and the
per-problem results of all 28 evaluation runs.
Runs of the proprietary model require AWS Bedrock access; the open-weight runs reproduce
with vLLM on a single multi-GPU machine.
\fi

\ificlrfinal
\subsubsection*{Acknowledgments}
We gratefully acknowledge the Laude Institute for generously providing the compute
resources that made this research possible. We also thank Prof.\ Tatsunori Hashimoto,
Prof.\ Michael Genesereth, and Dr.\ Simon Rubinstein-Salzedo for their insightful feedback
and support.
\fi

\bibliographystyle{iclr2027/iclr2027_conference}
\bibliography{references}

\begin{thebibliography}{70}
\providecommand{\natexlab}[1]{#1}
\providecommand{\url}[1]{\texttt{#1}}
\expandafter\ifx\csname urlstyle\endcsname\relax
  \providecommand{\doi}[1]{doi: #1}\else
  \providecommand{\doi}{doi: \begingroup \urlstyle{rm}\Url}\fi

\bibitem[Achim et~al.(2025)Achim, Best, Bietti, Der, F{\'e}d{\'e}rico, Gukov,
  Halpern-Leistner, Henningsgard, Kudryashov, Meiburg,
  et~al.]{achim2025aristotle}
Tudor Achim, Alex Best, Alberto Bietti, Kevin Der, Math{\"\i}s
  F{\'e}d{\'e}rico, Sergei Gukov, Daniel Halpern-Leistner, Kirsten
  Henningsgard, Yury Kudryashov, Alexander Meiburg, et~al.
\newblock Aristotle: {IMO}-level automated theorem proving.
\newblock \emph{arXiv preprint arXiv:2510.01346}, 2025.

\bibitem[Alon et~al.(2026)Alon, Bloom, Gowers, Litt, Sawin, Shankar, Tsimerman,
  Wang, and Wood]{alon2026remarks}
Noga Alon, Thomas~F. Bloom, W.~Timothy Gowers, Daniel Litt, Will Sawin, Arul
  Shankar, Jacob Tsimerman, Victor Wang, and Melanie~Matchett Wood.
\newblock Remarks on the disproof of the unit distance conjecture.
\newblock \emph{arXiv preprint arXiv:2605.20695}, 2026.

\bibitem[{Anthropic}(2026)]{anthropic2026sonnet46}
{Anthropic}.
\newblock System card: {Claude Sonnet 4.6}.
\newblock \url{https://www.anthropic.com/claude-sonnet-4-6-system-card}, 2026.

\bibitem[Ataeva et~al.(2026)Ataeva, Khalov, and Tuchkova]{ataeva2026multilevel}
Olga~M. Ataeva, Andrey~P. Khalov, and Natalia~P. Tuchkova.
\newblock Multi-level knowledge graph for formal mathematics.
\newblock \emph{Lobachevskii Journal of Mathematics}, 2026.
\newblock Accepted April 22, 2026.
  \url{https://openreview.net/forum?id=eZqzdol80P}.

\bibitem[Avigad et~al.(2026)Avigad, de~Moura, Kong, and
  Ullrich]{avigad2026tpil}
Jeremy Avigad, Leonardo de~Moura, Soonho Kong, and Sebastian Ullrich.
\newblock Theorem proving in {Lean 4}.
\newblock \url{https://lean-lang.org/theorem_proving_in_lean4/}, 2026.
\newblock Accessed September 18, 2026.

\bibitem[Azerbayev et~al.(2023)Azerbayev, Piotrowski, Schoelkopf, Ayers, Radev,
  and Avigad]{azerbayev2023proofnet}
Zhangir Azerbayev, Bartosz Piotrowski, Hailey Schoelkopf, Edward~W Ayers,
  Dragomir Radev, and Jeremy Avigad.
\newblock {ProofNet}: Autoformalizing and formally proving undergraduate-level
  mathematics.
\newblock \emph{arXiv preprint arXiv:2302.12433}, 2023.

\bibitem[Azerbayev et~al.(2024)Azerbayev, Schoelkopf, Paster, Dos~Santos,
  McAleer, Jiang, Deng, Biderman, and Welleck]{azerbayev2023llemma}
Zhangir Azerbayev, Hailey Schoelkopf, Keiran Paster, Marco Dos~Santos, Stephen
  McAleer, Albert~Q. Jiang, Jia Deng, Stella Biderman, and Sean Welleck.
\newblock Llemma: An open language model for mathematics.
\newblock In \emph{International Conference on Learning Representations},
  volume 2024, pp.\  40622--40649, 2024.

\bibitem[Bailey et~al.(2026)Bailey, Wen, Dong, Hashimoto, and
  Ma]{bailey2026sgs}
Luke Bailey, Kaiyue Wen, Kefan Dong, Tatsunori Hashimoto, and Tengyu Ma.
\newblock Scaling self-play with self-guidance.
\newblock \emph{arXiv preprint arXiv:2604.20209}, 2026.

\bibitem[Barkallah et~al.(2026)Barkallah, Bailey, Wen, Abouzaid, and
  Ma]{barkallah2026pseudoformalization}
Slim Barkallah, Luke Bailey, Kaiyue Wen, Mohammed Abouzaid, and Tengyu Ma.
\newblock {Pseudo-Formalization} for automatic proof verification.
\newblock \emph{arXiv preprint arXiv:2605.20531}, 2026.

\bibitem[Bian et~al.(2025)Bian, Geng, Yang, and Cheng]{zhang2025automathkg}
Rong Bian, Yu~Geng, Zijian Yang, and Bing Cheng.
\newblock {AutoMathKG}: The automated mathematical knowledge graph based on
  {LLM} and vector database.
\newblock \emph{Computational Intelligence}, 41\penalty0 (4):\penalty0 e70096,
  2025.

\bibitem[Chen et~al.(2025{\natexlab{a}})Chen, Chen, Du, Hu, Jiang, Jie, Jin,
  Jin, Li, Shi, et~al.]{chen2025seedprover15}
Jiangjie Chen, Wenxiang Chen, Jiacheng Du, Jinyi Hu, Zhicheng Jiang, Allan Jie,
  Xiaoran Jin, Xing Jin, Chenggang Li, Wenlei Shi, et~al.
\newblock {Seed-Prover} 1.5: Mastering undergraduate-level theorem proving via
  learning from experience.
\newblock \emph{arXiv preprint arXiv:2512.17260}, 2025{\natexlab{a}}.

\bibitem[Chen et~al.(2025{\natexlab{b}})Chen, Gu, Huang, Huang, Jiang, Jie,
  Jin, Jin, Li, Ma, et~al.]{chen2025seedprover}
Luoxin Chen, Jinming Gu, Liankai Huang, Wenhao Huang, Zhicheng Jiang, Allan
  Jie, Xiaoran Jin, Xing Jin, Chenggang Li, Kaijing Ma, et~al.
\newblock {Seed-Prover}: Deep and broad reasoning for automated theorem
  proving.
\newblock \emph{arXiv preprint arXiv:2507.23726}, 2025{\natexlab{b}}.

\bibitem[Curry(1934)]{curry1934functionality}
Haskell~B. Curry.
\newblock Functionality in combinatory logic.
\newblock \emph{Proceedings of the National Academy of Sciences}, 20\penalty0
  (11):\penalty0 584--590, 1934.

\bibitem[Dasgupta(2026)]{dasgupta2026geometric}
Ashani Dasgupta.
\newblock Geometric structure in the {Lean} 4 {Mathlib} knowledge graph, 2026.
\newblock Manuscript.
  \url{https://faculty.tcu.edu/gfriedman/gtw_archive/Problems/Dasgupta.pdf}.

\bibitem[Dong \& Ma(2025)Dong and Ma]{dong2025stp}
Kefan Dong and Tengyu Ma.
\newblock {STP}: Self-play {LLM} theorem provers with iterative conjecturing
  and proving.
\newblock \emph{arXiv preprint arXiv:2502.00212}, 2025.

\bibitem[Gao et~al.(2026)Gao, Sun, Jiang, Wang, Xu, Wu, Dai, and
  Dong]{gao2026leansearch}
Guoxiong Gao, Zeming Sun, Jiedong Jiang, Yutong Wang, Jingda Xu, Peihao Wu,
  Bryan Dai, and Bin Dong.
\newblock {LeanSearch} v2: Global premise retrieval for {Lean} 4 theorem
  proving.
\newblock \emph{arXiv preprint arXiv:2605.13137}, 2026.

\bibitem[Georgiev et~al.(2025)Georgiev, G{\'o}mez-Serrano, Tao, and
  Wagner]{georgiev2025exploration}
Bogdan Georgiev, Javier G{\'o}mez-Serrano, Terence Tao, and Adam~Zsolt Wagner.
\newblock Mathematical exploration and discovery at scale.
\newblock \emph{arXiv preprint arXiv:2511.02864}, 2025.

\bibitem[Hong et~al.(2024)Hong, Zhuge, Chen, Zheng, Cheng, Wang, Zhang, Yau,
  Lin, Zhou, et~al.]{hong2023metagpt}
Sirui Hong, Mingchen Zhuge, Jonathan Chen, Xiawu Zheng, Yuheng Cheng, Jinlin
  Wang, Ceyao Zhang, Steven Yau, Zijuan Lin, Liyang Zhou, et~al.
\newblock {MetaGPT}: Meta programming for a multi-agent collaborative
  framework.
\newblock In \emph{International Conference on Learning Representations},
  volume 2024, pp.\  23247--23275, 2024.

\bibitem[Howard(1980)]{howard1980formulae}
William~A. Howard.
\newblock The formulae-as-types notion of construction.
\newblock In Jonathan~P. Seldin and J.~Roger Hindley (eds.), \emph{To {H. B.
  Curry}: Essays on Combinatory Logic, Lambda Calculus and Formalism}, pp.\
  479--490. Academic Press, 1980.

\bibitem[Hubert et~al.(2026)Hubert, Mehta, Sartran, Horv{\'a}th,
  {\v{Z}}u{\v{z}}i{\'c}, Wieser, Huang, Schrittwieser, Schroecker, Masoom,
  et~al.]{hubert2026olympiad}
Thomas Hubert, Rishi Mehta, Laurent Sartran, Mikl{\'o}s~Z Horv{\'a}th, Goran
  {\v{Z}}u{\v{z}}i{\'c}, Eric Wieser, Aja Huang, Julian Schrittwieser, Yannick
  Schroecker, Hussain Masoom, et~al.
\newblock Olympiad-level formal mathematical reasoning with reinforcement
  learning.
\newblock \emph{Nature}, 651\penalty0 (8106):\penalty0 607--613, 2026.

\bibitem[Irving et~al.(2016)Irving, Szegedy, Alemi, Een, Chollet, and
  Urban]{irving2016deepmath}
Geoffrey Irving, Christian Szegedy, Alexander~A Alemi, Niklas Een, Francois
  Chollet, and Josef Urban.
\newblock Deepmath - deep sequence models for premise selection.
\newblock In \emph{Advances in Neural Information Processing Systems},
  volume~29, 2016.

\bibitem[Jiang et~al.(2022)Jiang, Welleck, Zhou, Li, Liu, Jamnik, Lacroix, Wu,
  and Lample]{jiang2022draftsketchprove}
Albert~Q Jiang, Sean Welleck, Jin~Peng Zhou, Wenda Li, Jiacheng Liu, Mateja
  Jamnik, Timoth{\'e}e Lacroix, Yuhuai Wu, and Guillaume Lample.
\newblock Draft, sketch, and prove: Guiding formal theorem provers with
  informal proofs.
\newblock \emph{arXiv preprint arXiv:2210.12283}, 2022.

\bibitem[Kabra et~al.(2026)Kabra, Jiang, Zhao, Garbaya, Mukherjee, Rute, Unsal,
  Soletskyi, and Sorg]{kabra2026leanstral}
Aditi Kabra, Albert~Q. Jiang, Andrew Zhao, Dhia Garbaya, Indraneel Mukherjee,
  Jason Rute, Mert Unsal, Roman Soletskyi, and Simon Sorg.
\newblock {Leanstral}, 2026.
\newblock URL \url{https://www.alphaxiv.org/abs/2608.leanstral}.

\bibitem[Kerger(2026)]{kerger2026matek}
Phillip Kerger.
\newblock {MATEK}: An open orchestration framework for agentic mathematical
  research with persistent knowledge graphs.
\newblock 2026.
\newblock SSRN preprint,
  \url{https://papers.ssrn.com/sol3/papers.cfm?abstract_id=7163458}.

\bibitem[Khalov et~al.(2026)Khalov, Ataeva, and
  Tuchkova]{khalov2026explainable}
Andrey Khalov, Olga Ataeva, and Natalia Tuchkova.
\newblock Explainable {AI} for mathematics: Proofs as code with knowledge graph
  and domain ontology support.
\newblock In \emph{Conference of Mathematics of AI}, 2026.
\newblock
  \url{https://openreview.net/pdf/775c1617ab35a9644ff0eb8310a9579427a3c88e.pdf}.

\bibitem[Kung et~al.(2026)Kung, Song, Hwang, Yoon, Li, Severini,
  Ol{\v{s}}{\'a}k, Lockhart, Le, Gokturk, et~al.]{kung2026leap}
Po-Nien Kung, Linfeng Song, Dawsen Hwang, Jinsung Yoon, Chun-Liang Li, Simone
  Severini, Mirek Ol{\v{s}}{\'a}k, Edward Lockhart, Quoc~V Le, Burak Gokturk,
  et~al.
\newblock {LEAP}: Supercharging {LLMs} for formal mathematics with agentic
  frameworks.
\newblock \emph{arXiv preprint arXiv:2606.03303}, 2026.

\bibitem[Kurgan et~al.(2026)Kurgan, Wang, Leonen, Szeto, Alexander, Remizov,
  Alper, Inchiostro, and Ilin]{kurgan2026theoremgraph}
Simon Kurgan, Evan Wang, Eric Leonen, Sophie Szeto, Luke Alexander, Artemii
  Remizov, Jarod Alper, Giovanni Inchiostro, and Vasily Ilin.
\newblock {TheoremGraph}: Bridging formal and informal mathematics.
\newblock \emph{arXiv preprint arXiv:2606.25363}, 2026.

\bibitem[Lample et~al.(2022)Lample, Lacroix, Lachaux, Rodriguez, Hayat, Lavril,
  Ebner, and Martinet]{lample2022htps}
Guillaume Lample, Timothee Lacroix, Marie-Anne Lachaux, Aurelien Rodriguez,
  Amaury Hayat, Thibaut Lavril, Gabriel Ebner, and Xavier Martinet.
\newblock Hypertree proof search for neural theorem proving.
\newblock \emph{Advances in neural information processing systems},
  35:\penalty0 26337--26349, 2022.

\bibitem[Leang et~al.(2026)Leang, Zhao, Stoian, Xu, Li, Li, Cohen, and
  Giunchiglia]{leang2026pythagoras}
Joshua Ong~Jun Leang, Zheng Zhao, Mihaela~C{\u{a}}t{\u{a}}lina Stoian, Qiyuan
  Xu, Haonan Li, Wenda Li, Shay~B. Cohen, and Eleonora Giunchiglia.
\newblock {Pythagoras-Prover}: Advancing efficient formal proving via augmented
  {Lean} formalisation.
\newblock \emph{arXiv preprint arXiv:2606.12594}, 2026.

\bibitem[Lewkowycz et~al.(2022)Lewkowycz, Andreassen, Dohan, Dyer, Michalewski,
  Ramasesh, Slone, Anil, Schlag, Gutman-Solo, et~al.]{lewkowycz2022minerva}
Aitor Lewkowycz, Anders Andreassen, David Dohan, Ethan Dyer, Henryk
  Michalewski, Vinay Ramasesh, Ambrose Slone, Cem Anil, Imanol Schlag, Theo
  Gutman-Solo, et~al.
\newblock Solving quantitative reasoning problems with language models.
\newblock \emph{Advances in Neural Information Processing Systems},
  35:\penalty0 3843--3857, 2022.

\bibitem[Li et~al.(2025)Li, Knappe, Fu, Han, and Zhu]{wang2025kgprover}
Vincent Li, Tim Knappe, Yule Fu, Kevin Han, and Kevin Zhu.
\newblock Scaling natural-language graph-based test time compute for automated
  theorem proving.
\newblock \emph{arXiv preprint arXiv:2503.11657}, 2025.

\bibitem[Li et~al.(2026)Li, Peng, Severini, and Shafto]{li2026network}
Xinze Li, Nanyun Peng, Simone Severini, and Patrick Shafto.
\newblock The network structure of {Mathlib}.
\newblock \emph{arXiv preprint arXiv:2604.24797}, 2026.

\bibitem[Lin et~al.(2025)Lin, Tang, Lyu, Wu, Lin, Yang, Li, Xia, Chen, Arora,
  and Jin]{lin2025goedelprover}
Yong Lin, Shange Tang, Bohan Lyu, Jiayun Wu, Hongzhou Lin, Kaiyu Yang, Jia Li,
  Mengzhou Xia, Danqi Chen, Sanjeev Arora, and Chi Jin.
\newblock {Goedel-Prover}: A frontier model for open-source automated theorem
  proving.
\newblock \emph{arXiv preprint arXiv:2502.07640}, 2025.

\bibitem[Lin et~al.(2026)Lin, Tang, Lyu, Yang, Chung, Zhao, Jiang, Geng, Ge,
  Sun, et~al.]{lin2026goedel}
Yong Lin, Shange Tang, Bohan Lyu, Ziran Yang, Jui-Hui Chung, Haoyu Zhao, Lai
  Jiang, Yihan Geng, Jiawei Ge, Jingruo Sun, et~al.
\newblock {Goedel-Prover-V2}: Scaling formal theorem proving with scaffolded
  data synthesis and self-correction.
\newblock In \emph{International Conference on Learning Representations},
  volume 2026, pp.\  11793--11818, 2026.

\bibitem[Luo et~al.(2025)Luo, Sun, Xu, Zhao, Lou, Tao, Geng, Lin, Chen, Tang,
  et~al.]{luo2025wizardmath}
Haipeng Luo, Qingfeng Sun, Can Xu, Pu~Zhao, Jian-Guang Lou, Chongyang Tao,
  Xiubo Geng, Qingwei Lin, Shifeng Chen, Yansong Tang, et~al.
\newblock {WizardMath}: Empowering mathematical reasoning for large language
  models via reinforced evol-instruct.
\newblock In \emph{International Conference on Learning Representations},
  volume 2025, pp.\  49573--49609, 2025.

\bibitem[Mallen et~al.(2023)Mallen, Asai, Zhong, Das, Khashabi, and
  Hajishirzi]{mallen2023trust}
Alex Mallen, Akari Asai, Victor Zhong, Rajarshi Das, Daniel Khashabi, and
  Hannaneh Hajishirzi.
\newblock When not to trust language models: Investigating effectiveness of
  parametric and non-parametric memories.
\newblock In \emph{Proceedings of the 61st annual meeting of the association
  for computational linguistics (volume 1: Long papers)}, pp.\  9802--9822,
  2023.

\bibitem[Miku{\l}a et~al.(2024)Miku{\l}a, Tworkowski, Antoniak, Piotrowski,
  Jiang, Zhou, Szegedy, Kuci{\'n}ski, Mi{\l}o{\'s}, and
  Wu]{mikula2024magnushammer}
Maciej Miku{\l}a, Szymon Tworkowski, Szymon Antoniak, Bartosz Piotrowski,
  Qiaochu Jiang, Jin Zhou, Christian Szegedy, {\L}ukasz Kuci{\'n}ski, Piotr
  Mi{\l}o{\'s}, and Yuhuai Wu.
\newblock Magnushammer: A transformer-based approach to premise selection.
\newblock In \emph{International Conference on Learning Representations},
  volume 2024, pp.\  39326--39350, 2024.

\bibitem[Moura \& Ullrich(2021)Moura and Ullrich]{moura2021lean4}
Leonardo~de Moura and Sebastian Ullrich.
\newblock The lean 4 theorem prover and programming language.
\newblock In \emph{International Conference on Automated Deduction}, pp.\
  625--635. Springer, 2021.

\bibitem[Novikov et~al.(2025)Novikov, V{\~u}, Eisenberger, Dupont, Huang,
  Wagner, Shirobokov, Kozlovskii, Ruiz, Mehrabian,
  et~al.]{novikov2025alphaevolve}
Alexander Novikov, Ng{\^a}n V{\~u}, Marvin Eisenberger, Emilien Dupont, Po-Sen
  Huang, Adam~Zsolt Wagner, Sergey Shirobokov, Borislav Kozlovskii,
  Francisco~JR Ruiz, Abbas Mehrabian, et~al.
\newblock {AlphaEvolve}: A coding agent for scientific and algorithmic
  discovery.
\newblock \emph{arXiv preprint arXiv:2506.13131}, 2025.

\bibitem[Ono(2026)]{ono2026arctan}
Ken Ono.
\newblock Integer values of $\tan(\arctan 1+\arctan 2+\cdots+\arctan n)$ are
  rare.
\newblock \emph{arXiv preprint arXiv:2607.05739}, 2026.

\bibitem[OpenAI(2026)]{openai2026tenadvances}
OpenAI.
\newblock Ten advances in mathematics and theoretical computer science, 2026.
\newblock URL \url{https://www.alphaxiv.org/abs/2607.ten-advancements}.

\bibitem[Ospanov et~al.(2026)Ospanov, Feng, Sun, Bai, Shen, and
  Farnia]{ospanov2026hermes}
Azim Ospanov, Zijin Feng, Jiacheng Sun, Haoli Bai, Xin Shen, and Farzan Farnia.
\newblock {HERMES}: Towards efficient and verifiable mathematical reasoning in
  {LLMs}.
\newblock In \emph{Forty-third International Conference on Machine Learning},
  2026.
\newblock URL \url{https://openreview.net/forum?id=w7BZcUc0fJ}.

\bibitem[Pham et~al.(2026)Pham, Karimov, Galichin, and
  Oseledets]{pham2026theorembench}
QuocViet Pham, Elvir Karimov, Andrey Galichin, and Ivan Oseledets.
\newblock {TheoremBench}: Evaluating {LLMs} on theorem proving in formal
  mathematics.
\newblock \emph{arXiv preprint arXiv:2606.09450}, 2026.

\bibitem[Polu \& Sutskever(2020)Polu and Sutskever]{polu2020gptf}
Stanislas Polu and Ilya Sutskever.
\newblock Generative language modeling for automated theorem proving.
\newblock \emph{arXiv preprint arXiv:2009.03393}, 2020.

\bibitem[Rammal et~al.(2026)Rammal, Patel, Gloeckle, Hayat, Kempe, Munos,
  Arnal, and Cabannes]{rammal2026formalizing}
Ahmad Rammal, Niket Patel, Fabian Gloeckle, Amaury Hayat, Julia Kempe, Remi
  Munos, Charles Arnal, and Vivien Cabannes.
\newblock Formalizing mathematics at scale.
\newblock \emph{arXiv preprint arXiv:2605.29955}, 2026.

\bibitem[Ren et~al.(2025)Ren, Shao, Song, Xin, Wang, Zhao, Zhang, Fu, Zhu,
  Yang, et~al.]{ren2025deepseekproverv2}
ZZ~Ren, Zhihong Shao, Junxiao Song, Huajian Xin, Haocheng Wang, Wanjia Zhao,
  Liyue Zhang, Zhe Fu, Qihao Zhu, Dejian Yang, et~al.
\newblock {DeepSeek-Prover-V2}: Advancing formal mathematical reasoning via
  reinforcement learning for subgoal decomposition.
\newblock \emph{arXiv preprint arXiv:2504.21801}, 2025.

\bibitem[Romera-Paredes et~al.(2024)Romera-Paredes, Barekatain, Novikov, Balog,
  Kumar, Dupont, Ruiz, Ellenberg, Wang, Fawzi, et~al.]{romera2024funsearch}
Bernardino Romera-Paredes, Mohammadamin Barekatain, Alexander Novikov, Matej
  Balog, M~Pawan Kumar, Emilien Dupont, Francisco~JR Ruiz, Jordan~S Ellenberg,
  Pengming Wang, Omar Fawzi, et~al.
\newblock Mathematical discoveries from program search with large language
  models.
\newblock \emph{Nature}, 625\penalty0 (7995):\penalty0 468--475, 2024.

\bibitem[Shen et~al.(2025)Shen, Huang, Yang, Wang, Gao, Xu, Jiang, He, Yang,
  Sun, Ju, Wu, Dai, and Dong]{shen2025realprover}
Ziju Shen, Naohao Huang, Fanyi Yang, Yutong Wang, Guoxiong Gao, Tianyi Xu,
  Jiedong Jiang, Wanyi He, Pu~Yang, Mengzhou Sun, Haocheng Ju, Peihao Wu, Bryan
  Dai, and Bin Dong.
\newblock {REAL-Prover}: Retrieval augmented {Lean} prover for mathematical
  reasoning.
\newblock \emph{arXiv preprint arXiv:2505.20613}, 2025.

\bibitem[Song et~al.(2025)Song, Yang, and Anandkumar]{song2025leancopilot}
Peiyang Song, Kaiyu Yang, and Anima Anandkumar.
\newblock {Lean Copilot}: Large language models as copilots for theorem proving
  in {Lean}.
\newblock In \emph{Proceedings of the International Conference on
  Neuro-symbolic Systems}, volume 288 of \emph{Proceedings of Machine Learning
  Research}, pp.\  144--169. PMLR, 2025.

\bibitem[Tao(2026)]{tao2026mathematics}
Terence Tao.
\newblock Mathematics in the age of {AI}.
\newblock \emph{arXiv preprint arXiv:2608.16753}, 2026.

\bibitem[Thakur et~al.(2024)Thakur, Tsoukalas, Wen, Xin, and
  Chaudhuri]{thakur2024copra}
Amitayush Thakur, George Tsoukalas, Yeming Wen, Jimmy Xin, and Swarat
  Chaudhuri.
\newblock An in-context learning agent for formal theorem-proving.
\newblock In \emph{Proceedings of the Conference on Language Modeling (COLM)},
  2024.
\newblock URL \url{https://openreview.net/forum?id=V7HRrxXUhN}.

\bibitem[{The mathlib Community}(2020)]{mathlib2020}
{The mathlib Community}.
\newblock The {Lean} mathematical library.
\newblock In \emph{Proceedings of the 9th {ACM} {SIGPLAN} International
  Conference on Certified Programs and Proofs}, CPP '20, pp.\  367--381. ACM,
  2020.
\newblock \doi{10.1145/3372885.3373824}.
\newblock URL \url{https://doi.org/10.1145/3372885.3373824}.

\bibitem[{The mathlib Community}(2026)]{mathlib_stats2026}
{The mathlib Community}.
\newblock {Mathlib} statistics.
\newblock \url{https://leanprover-community.github.io/mathlib_stats.html},
  2026.
\newblock Accessed September 13, 2026.

\bibitem[Tsoukalas et~al.(2024)Tsoukalas, Lee, Jennings, Xin, Ding, Jennings,
  Thakur, and Chaudhuri]{tsoukalas2024putnambench}
George Tsoukalas, Jasper Lee, John Jennings, Jimmy Xin, Michelle Ding, Michael
  Jennings, Amitayush Thakur, and Swarat Chaudhuri.
\newblock {PutnamBench}: Evaluating neural theorem-provers on the {Putnam
  Mathematical Competition}.
\newblock \emph{Advances in Neural Information Processing Systems},
  37:\penalty0 11545--11569, 2024.

\bibitem[Tsoukalas et~al.(2026)Tsoukalas, Kovsharov, Shirobokov, Surina,
  Firsching, B{\'e}rczi, Ruiz, Suggala, Wagner, Wieser,
  et~al.]{tsoukalas2026advancing}
George Tsoukalas, Anton Kovsharov, Sergey Shirobokov, Anja Surina, Moritz
  Firsching, Gergely B{\'e}rczi, Francisco~JR Ruiz, Arun Suggala, Adam~Zsolt
  Wagner, Eric Wieser, et~al.
\newblock Advancing mathematics research with {AI}-driven formal proof search.
\newblock \emph{arXiv preprint arXiv:2605.22763}, 2026.

\bibitem[Varambally et~al.(2026)Varambally, Voice, Sun, Chen, Yu, and
  Ye]{chen2025hilbert}
Sumanth Varambally, Thomas Voice, Yanchao Sun, Zhifeng Chen, Rose Yu, and
  Ke~Ye.
\newblock Hilbert: Recursively building formal proofs with informal reasoning.
\newblock In \emph{International Conference on Learning Representations},
  volume 2026, pp.\  38998--39031, 2026.

\bibitem[Wadhwa et~al.(2023)Wadhwa, Amir, and Wallace]{wadhwa2023revisiting}
Somin Wadhwa, Silvio Amir, and Byron~C Wallace.
\newblock Revisiting relation extraction in the era of large language models.
\newblock In \emph{Proceedings of the 61st Annual Meeting of the Association
  for Computational Linguistics}, pp.\  15566--15589, 2023.

\bibitem[Wang et~al.(2025)Wang, Unsal, Lin, Baksys, Liu, Santos, Sung, Vinyes,
  Ying, Zhu, et~al.]{sun2025kimina}
Haiming Wang, Mert Unsal, Xiaohan Lin, Mantas Baksys, Junqi Liu, Marco~Dos
  Santos, Flood Sung, Marina Vinyes, Zhenzhe Ying, Zekai Zhu, et~al.
\newblock {Kimina-Prover} preview: Towards large formal reasoning models with
  reinforcement learning.
\newblock \emph{arXiv preprint arXiv:2504.11354}, 2025.

\bibitem[Wang et~al.(2017)Wang, Tang, Wang, and Deng]{wang2017premise}
Mingzhe Wang, Yihe Tang, Jian Wang, and Jia Deng.
\newblock Premise selection for theorem proving by deep graph embedding.
\newblock In \emph{Advances in Neural Information Processing Systems},
  volume~30, 2017.

\bibitem[Welleck et~al.(2021)Welleck, Liu, Bras, Hajishirzi, Choi, and
  Cho]{welleck2021naturalproofs}
Sean Welleck, Jiacheng Liu, Ronan~Le Bras, Hannaneh Hajishirzi, Yejin Choi, and
  Kyunghyun Cho.
\newblock {NaturalProofs}: Mathematical theorem proving in natural language.
\newblock In \emph{Thirty-fifth Conference on Neural Information Processing
  Systems Datasets and Benchmarks Track}, 2021.
\newblock URL \url{https://openreview.net/forum?id=Jvxa8adr3iY}.

\bibitem[Wu et~al.(2024)Wu, Barrett, and Narodytska]{wu2024lemur}
Haoze Wu, Clark Barrett, and Nina Narodytska.
\newblock Lemur: Integrating large language models in automated program
  verification.
\newblock In \emph{International Conference on Learning Representations},
  volume 2024, pp.\  2968--2978, 2024.

\bibitem[Wu et~al.(2023)Wu, Bansal, Zhang, Wu, Li, Zhu, Jiang, Zhang, Zhang,
  Liu, et~al.]{wu2023autogen}
Qingyun Wu, Gagan Bansal, Jieyu Zhang, Yiran Wu, Beibin Li, Erkang Zhu,
  Li~Jiang, Xiaoyun Zhang, Shaokun Zhang, Jiale Liu, et~al.
\newblock {AutoGen}: Enabling next-gen {LLM} applications via multi-agent
  conversation.
\newblock \emph{arXiv preprint arXiv:2308.08155}, 2023.

\bibitem[Wu et~al.(2022)Wu, Jiang, Li, Rabe, Staats, Jamnik, and
  Szegedy]{wu2022autoformalization}
Yuhuai Wu, Albert~Qiaochu Jiang, Wenda Li, Markus Rabe, Charles Staats, Mateja
  Jamnik, and Christian Szegedy.
\newblock Autoformalization with large language models.
\newblock \emph{Advances in neural information processing systems},
  35:\penalty0 32353--32368, 2022.

\bibitem[Xin et~al.(2025)Xin, Zheng, Nie, Yuan, and Xiao]{liu2025bfsproverv2}
Ran Xin, Zeyu Zheng, Yanchen Nie, Kun Yuan, and Xia Xiao.
\newblock Scaling up multi-turn off-policy {RL} and multi-agent tree search for
  {LLM} step-provers.
\newblock \emph{arXiv preprint arXiv:2509.06493}, 2025.

\bibitem[Xu et~al.(2024)Xu, Chen, Peng, Zhang, Xu, Zhao, Wu, Zheng, Wang, and
  Chen]{xu2024generativeie}
Derong Xu, Wei Chen, Wenjun Peng, Chao Zhang, Tong Xu, Xiangyu Zhao, Xian Wu,
  Yefeng Zheng, Yang Wang, and Enhong Chen.
\newblock Large language models for generative information extraction: A
  survey.
\newblock \emph{Frontiers of Computer Science}, 18\penalty0 (6):\penalty0
  186357, 2024.

\bibitem[Yang et~al.(2025)Yang, Li, Yang, Zhang, Hui, Zheng, Yu, Gao, Huang,
  Lv, et~al.]{qwen3_2025}
An~Yang, Anfeng Li, Baosong Yang, Beichen Zhang, Binyuan Hui, Bo~Zheng, Bowen
  Yu, Chang Gao, Chengen Huang, Chenxu Lv, et~al.
\newblock {Qwen3} technical report.
\newblock \emph{arXiv preprint arXiv:2505.09388}, 2025.

\bibitem[Yang et~al.(2023)Yang, Swope, Gu, Chalamala, Song, Yu, Godil, Prenger,
  and Anandkumar]{yang2023leandojo}
Kaiyu Yang, Aidan Swope, Alex Gu, Rahul Chalamala, Peiyang Song, Shixing Yu,
  Saad Godil, Ryan~J Prenger, and Animashree Anandkumar.
\newblock {LeanDojo}: Theorem proving with retrieval-augmented language models.
\newblock \emph{Advances in Neural Information Processing Systems},
  36:\penalty0 21573--21612, 2023.

\bibitem[Ye et~al.(2022)Ye, Zhang, Chen, and Chen]{ye2022generative}
Hongbin Ye, Ningyu Zhang, Hui Chen, and Huajun Chen.
\newblock Generative knowledge graph construction: A review.
\newblock In \emph{Proceedings of the 2022 Conference on Empirical Methods in
  Natural Language Processing}, pp.\  1--17, 2022.

\bibitem[Zhang et~al.(2026)Zhang, Borchert, Gritta, and Lampouras]{li2025drift}
Meiru Zhang, Philipp Borchert, Milan Gritta, and Gerasimos Lampouras.
\newblock Drift: decompose, retrieve, illustrate, then formalize theorems.
\newblock In \emph{International Conference on Learning Representations},
  volume 2026, 2026.
\newblock URL \url{https://openreview.net/forum?id=IGEs577RFz}.

\bibitem[Zheng et~al.(2022)Zheng, Han, and Polu]{zheng2022minif2f}
Kunhao Zheng, Jesse~Michael Han, and Stanislas Polu.
\newblock {miniF2F}: A cross-system benchmark for formal olympiad-level
  mathematics.
\newblock In \emph{International Conference on Learning Representations}, 2022.
\newblock URL \url{https://openreview.net/forum?id=9ZPegFuFTFv}.

\end{thebibliography}

\clearpage
\appendix
\raggedbottom

\section{Background: Mathlib and Lean 4}
\label{app:mathlib_primer}

We briefly review the Lean~4 and \Mathlib{} concepts used in this paper.

\paragraph{Lean~4.}
Lean~4~\citep{moura2021lean4} is an interactive theorem prover and programming language;
\citet{avigad2026tpil} give a tutorial introduction.
Mathematical statements are expressed as \emph{types}, and proofs are \emph{terms} of
those types (the Curry--Howard correspondence;
\citealp{curry1934functionality,howard1980formulae}). For example, a proof of
$\forall n : \mathbb{N},\; 0 \leq n$ is a function that takes any natural number and
returns evidence of the inequality.

\paragraph{Mathlib.}
\Mathlib{}~\citep{mathlib2020} is the community-maintained mathematical library for
Lean~4, containing 285{,}000+ theorems and 135{,}000+ definitions as of September
2026~\citep{mathlib_stats2026}, spanning undergraduate through research-level mathematics.
Each declaration
is machine-verified: the Lean kernel checks every proof step.

\paragraph{Declaration kinds.}
The \Mathlib{} library contains several kinds of declarations:
\begin{itemize}
  \item \texttt{theorem}: A proved mathematical statement (e.g., Fermat's Little Theorem)
  \item \texttt{def}: A mathematical definition (e.g., \texttt{Nat.Prime}, \texttt{Nat.gcd})
  \item \texttt{class}: A typeclass (e.g., \texttt{Group}, \texttt{Ring}, \texttt{Field})
  \item \texttt{instance}: A typeclass instance (e.g., \texttt{Int} is a \texttt{Ring})
  \item \texttt{structure}: A data structure definition
  \item \texttt{inductive}: An inductive type definition
\end{itemize}

\paragraph{Type signatures.}
Every declaration has a \emph{type signature} that formally specifies what it states
or computes. For example, Fermat's Little Theorem has the type:

\begin{codebox}{Type signature: \texttt{ZMod.pow\_card\_sub\_one\_eq\_one} (Fermat's Little Theorem)}theorem ZMod.pow_card_sub_one_eq_one
  {p : @$\mathbb{N}$@} [Fact (Nat.Prime p)] {a : ZMod p} (ha : a @$\neq$@ 0) :
  a ^ (p - 1) = 1
\end{codebox}

\noindent Reading this signature: \texttt{p} is a natural number assumed to be prime
(via the \texttt{[Fact (Nat.Prime p)]} typeclass constraint); \texttt{a} is an element of
$\mathbb{Z}/p\mathbb{Z}$; \texttt{ha} is a proof that $a \neq 0$; and the conclusion
is $a^{p-1} = 1$. Lean's \texttt{exact?} tactic searches the imported environment by
\emph{type unification}: given a proof goal (a type), it
finds theorems whose conclusion type matches after substituting variables.

\paragraph{Declaration metadata.}
The \Mathlib{} documentation index (404{,}440 entries, covering \Mathlib{} and its
dependencies such as Lean core and Batteries) provides, for each declaration: (1)~the declaration
name (e.g., \texttt{ZMod.pow\_card\_sub\_one\_eq\_one}), (2)~its kind
(\texttt{theorem}/\texttt{def}/etc.), (3)~its full type signature, and (4)~its docstring;
proofs themselves are in the source files. Our Layer~2 search (\S\ref{sec:agents})
prefix-matches the extracted search terms against these declaration names and attaches the
matching signatures and docstrings (Appendix~\ref{app:explorer}).

\section{\MathKG{} Example Nodes}
\label{app:mathkg_node}

We show two \MathKG{} node entries, Sperner's Theorem and the Cayley--Hamilton Theorem, each
followed by the Lean~4 type signature that the
\Mathlib{} declaration database returns for its \texttt{lean4\_mathlib\_name}; the node
file itself stores the nine fields shown.

\begin{codebox}{Node record: Sperner's Theorem (Combinatorics)}
id: "thm_comb_004"
name: "Sperner's Theorem"
domain: "Combinatorics"
statement: "Max antichain in P([n]) has size
  C(n, @$\lfloor n/2 \rfloor$@)."
type: "theorem"
keywords: ["Sperner", "antichain", "extremal"]
lean4_mathlib_name: "IsAntichain.sperner"
lean4_mathlib_import: "Mathlib.Combinatorics.SetFamily.LYM"
lean4_status: "verified"
\end{codebox}
\begin{codebox}{Retrieved Lean~4 type signature: \texttt{IsAntichain.sperner}}
theorem IsAntichain.sperner {@$\alpha$@ : Type u_2} [Fintype @$\alpha$@]
  {@$\mathcal{A}$@ : Finset (Finset @$\alpha$@)}
  (h@$\mathcal{A}$@ : IsAntichain (fun (x1 x2 : Finset @$\alpha$@) => x1 @$\subseteq$@ x2) @$\uparrow\mathcal{A}$@) :
  @$\mathcal{A}$@.card @$\leq$@ (Fintype.card @$\alpha$@).choose (Fintype.card @$\alpha$@ / 2)
\end{codebox}

\begin{codebox}{Node record: Cayley--Hamilton Theorem (Linear Algebra)}
id: "thm_la_002"
name: "Cayley-Hamilton Theorem"
domain: "Linear Algebra"
statement: "Every square matrix satisfies its own
  characteristic polynomial."
type: "theorem"
keywords: ["Cayley-Hamilton", "characteristic polynomial"]
lean4_mathlib_name: "Matrix.aeval_self_charpoly"
lean4_mathlib_import: "Mathlib.LinearAlgebra.Matrix.Charpoly.Basic"
lean4_status: "verified"
\end{codebox}
\begin{codebox}{Retrieved Lean~4 type signature: \texttt{Matrix.aeval\_self\_charpoly}}
theorem Matrix.aeval_self_charpoly {R : Type u_1} [CommRing R]
  {n : Type u_4} [DecidableEq n] [Fintype n] (M : Matrix n n R) :
  (Polynomial.aeval M) M.charpoly = 0
\end{codebox}

\section{Explorer Search Procedure}
\label{app:explorer}

This appendix gives the full detail of the two-layer Explorer search summarized in
\S\ref{sec:agents}. In all reported runs the Explorer's LLM calls are answered by the
evaluated model itself, so each model's context reflects its own selection quality; a
stronger selector might raise the smaller models' gains.

\paragraph{Layer 1: \MathKG{} semantic search.} The Explorer searches \MathKG{} in four steps:
\begin{enumerate}
  \item \textbf{Domain pre-filter.} Restrict candidates to nodes whose domain label
        contains the problem's domain, plus their cross-domain-bridge neighbors. A number
        theory problem keeps 172 of the 364 nodes (49 number theory nodes plus 123 bridge
        neighbors); an algebra problem matches three domains (Algebra, Abstract Algebra,
        and Linear Algebra) and keeps 324. When no node
        domain matches the label, seed selection runs over the whole graph. On miniF2F,
        labels are inferred from problem identifiers; this applies to the 175 problems
        whose identifiers name a competition or a proof technique rather than a topic
        (``competition'': 120 AMC/AIME problems, ``imo'': 39, ``other'': 16 induction
        problems).
        For these problems, all 364 nodes are candidates for seed selection.
  \item \textbf{Seed selection.} Send each candidate's name, statement, and keywords to
        the evaluated model, which scores their relevance to the problem's formal statement
        and returns $s$ seeds (default $s{=}5$), each with a one-line rationale.
  \item \textbf{Edge walk.} From each seed, perform breadth-first traversal of the
        \MathKG{} edges in both directions up to a configurable depth (default 1 hop).
        For each discovered
        node, the edge \emph{reasoning text} (generated during edge inference) is
        preserved, explaining \emph{why} the connection exists.
  \item \textbf{Context assembly.} For every node in the result set, emit its name and
        statement, tagged \texttt{SEED} or \texttt{CONNECTED}; seeds carry the scorer's
        relevance rationale, and neighbors carry a
        connection note (edge type and originating seed) plus one line per connecting edge
        with the edge type, confidence, and the leading 100 characters of the edge
        reasoning, to bound the block's size. The block itself is
        uncapped: every one-hop neighbor of every seed is included.
\end{enumerate}
The output is a structured context block where seed nodes appear first (highest
relevance), followed by edge-connected neighbors sorted by the highest confidence among
their connecting edges.

\paragraph{Layer 2: \Mathlib{} declaration search.} The Explorer searches the
declaration names of the \Mathlib{} documentation index (\S\ref{sec:mathkg}) through a
four-step pipeline:
\begin{enumerate}
  \item \textbf{Search term extraction.} Two complementary sources provide search terms:
        (a)~regex extraction of Lean~4 identifiers directly from the formal statement
        (e.g., {\small\texttt{Nat.Prime}}, {\small\texttt{ZMod}}); (b)~an LLM call asks
        the evaluated model to suggest 10--15 \Mathlib{} namespace prefixes tailored to the
        problem (e.g., {\small\texttt{Function.Injective}},
        {\small\texttt{Nat.strongRecOn}}), each classified as \emph{essential},
        \emph{helpful}, or \emph{speculative}; the two sources are merged into one set of terms.
        The regex receives the full formal statement, and the LLM call receives the
        natural-language statement plus the first 500 characters of the formal statement.
  \item \textbf{Prefix matching.} Each term is matched, as an exact name or a namespace
        prefix, against the theorem and definition names of the \Mathlib{} documentation
        index (Appendix~\ref{app:mathlib_primer}). For example, \texttt{Nat.Prime} matches
        110 declarations.
  \item \textbf{Scoring and ranking.} Each matched declaration receives a composite score
        from five signals: \emph{source priority} (formal statement identifiers score
        highest; LLM essential $>$ helpful $>$ speculative), \emph{match precision} (exact
        name match $>$ prefix match), \emph{name depth} (fewer namespace levels $=$ more
        fundamental), \emph{declaration kind} (theorem $>$ definition), and \emph{module
        proximity} (declarations in modules mapped to the problem's topic label receive a
        1.2$\times$ boost; problems without a topic label receive no boost). Results are
        capped per search term (8 for essential, 5 for helpful, 2 for speculative).
  \item \textbf{Type-signature enrichment.} Each matched declaration is enriched with its
        full Lean~4 type signature and the first 200 characters of its docstring from the
        pre-processed database of 404{,}440 declarations. For example, instead of returning
        only the name {\small\texttt{ZMod.pow\_card\_sub\_one\_eq\_one}}, the Prover
        receives (docstring wrapped here):
        \begin{codebox}{Enriched Layer~2 result: \texttt{ZMod.pow\_card\_sub\_one\_eq\_one}}theorem ZMod.pow_card_sub_one_eq_one
  {p : @$\mathbb{N}$@} [Fact (Nat.Prime p)] {a : ZMod p}
  (ha : a @$\neq$@ 0) : a ^ (p - 1) = 1
  -- **Fermat's Little Theorem**: for all nonzero
  -- `a : ZMod p`, we have `a ^ (p - 1) = 1`.
        \end{codebox}
        This enables the Prover to (a) use exact names without hallucination,
        (b) understand what arguments each lemma requires, and (c) judge whether a lemma
        is tactically relevant to the current goal. Type strings are stored as rendered
        by the documentation site, so about 14\% of \Mathlib{} theorem and definition
        entries carry HTML-escaped operators (\texttt{\&lt;} and \texttt{\&gt;} for
        \texttt{<}, \texttt{>}, and \texttt{=>}). The same database and code served every
        model and both \Mathlib{} modes (\texttt{with\_mathlib} and \texttt{full\_system}),
        so all runs received these strings identically and the cross-model comparison is
        unaffected.
\end{enumerate}

\paragraph{Full system (Layers 1+2).} When both layers run together, the Layer~1 block is
emitted first under the heading \texttt{RELEVANT THEOREMS FROM KNOWLEDGE GRAPH}, followed
by the Layer~2 block under \texttt{MATHLIB LEMMAS (from library search)}. In this mode
Layer~2 is KG-aware: a declaration whose module matches the \Mathlib{} import of a Layer~1
seed receives an additional 1.5$\times$ boost, and declarations already present in the
Layer~1 block are omitted from the Layer~2 list.

\section{\MathKG{} Edge Construction}
\label{app:relation_extraction}

\subsection{Relation Extraction Prompt}
\label{app:edge_prompt}

Below is the complete, verbatim prompt sent to the edge-inference LLM (Claude Sonnet~4.6
via AWS Bedrock) for each directed pair $(A, B)$.  The pair shown is a concrete
instantiation for \texttt{thm\_comb\_003} (Binomial Theorem) as node~$A$ and
\texttt{thm\_comb\_027} (Sum of Binomial Coefficients) as node~$B$; for every other
call, only the two \texttt{NODE} blocks change: the ontology, few-shot examples,
and output schema are identical across all $132{,}132$ directed
pairs.  Its valid outputs are the eight types of Table~\ref{tab:edge_types} plus
\texttt{none}, matching the relation extraction pipeline
(\S\ref{sec:relation_extraction}).

\begin{promptbox}{Relation extraction prompt (Claude Sonnet 4.6, one directed pair)}
You are a mathematical reasoning expert building a knowledge graph of fundamental mathematics. You must classify the relationship from NODE A to NODE B (directional) using a precise ontology. Be conservative: only identify strong, clear relationships.

NODE A (THEOREM):
Name: Binomial Theorem
Domain: Combinatorics
Statement: (x+y)^n = ~$\Sigma$~ C(n,k) x^{n-k} y^k.
Keywords: binomial, expansion
Mathlib name: Commute.add_pow
Mathlib import: Mathlib.Data.Nat.Choose.Sum

NODE B (THEOREM):
Name: Sum of Binomial Coefficients
Domain: Combinatorics
Statement: ~$\Sigma$~_{k=0}^n C(n,k) = 2^n.
Keywords: binomial, sum
Mathlib name: Nat.sum_range_choose
Mathlib import: Mathlib.Data.Nat.Choose.Sum

==== ONTOLOGY (8 edge types) ====

1. prerequisite
   A (typically a DEFINITION) is required to state or prove B.
   Example: "Prime Number" --prerequisite--> "Fermat's Little Theorem"
   Restriction: A must be a DEFINITION. If A is a theorem used inside B's proof, use "applied-in" instead.

2. generalizes
   A is strictly more general; B is recoverable from A by fixing parameters/structure.
   Example: "Binomial Theorem" --generalizes--> "Sum of Binomial Coefficients = 2^n"
   Restriction: Both A and B must be THEOREMS (not definitions).

3. specializes
   Inverse of generalizes. A is a strict special case of B.
   Example: "Fermat's Little Theorem" --specializes--> "Lagrange's Theorem (group theory)"
   Restriction: Both A and B must be THEOREMS.

4. equivalent-to
   A ~$\leftrightarrow$~ B at the same level of abstraction. Different formulations of the same fact.
   Example: "Heine-Borel (compact iff closed and bounded in R^n)" --equivalent-to--> "Bolzano-Weierstrass (bounded sequence has convergent subsequence in R^n)"
   Restriction: Both A and B must be THEOREMS. Pick only when both directions genuinely hold. Do NOT use for "related" or "both about X".

5. applied-in
   A's RESULT is used as a step in B's proof.
   Example: "Fundamental Theorem of Arithmetic" --applied-in--> "Lifting the Exponent Lemma"
   Restriction: Both A and B must be THEOREMS. A is a tool; B is what uses the tool. Definitions do not have proofs, so they cannot appear as B. If A is a definition, use "prerequisite" instead.

6. analogous-to
   A and B have the same formal structure (same theorem schema or proof pattern) in different domains or contexts.
   Example: "Hensel's Lemma (lifting roots p-adically)" --analogous-to--> "Implicit Function Theorem (lifting solutions smoothly)"
   Restriction: Both A and B must be THEOREMS. Must be a real structural correspondence, not just "both involve X". If you cannot state the shared schema in one sentence, it is not analogous.

7. cross-domain-bridge
   A pair where A's domain and B's domain meet: use this when ANY of the following hold:
     (a) one side's object is defined using the other's structure, OR
     (b) a canonical/standard theorem establishes a formal correspondence between them
         (e.g., gcd(a,b)~$\cdot$~Z = (a,b) as an ideal ~\textrm{\textemdash}~ the formal link between GCD and Ideal), OR
     (c) translation between the two domains routinely passes through this pair.
   Example: "Vector Space" --cross-domain-bridge--> "Field" (vector spaces are modules over fields)
   Example: "Greatest Common Divisor" --cross-domain-bridge--> "Ideal" (gcd(a,b) generates the ideal (a,b) in Z)
   Restriction: Nodes must be in DIFFERENT domains. The connection must be a recognized correspondence,
   not mere surface similarity.

8. instance-of
   A is a concrete object that exemplifies the abstract structure B.
   Example: "Ring ZMod n" --instance-of--> "Ring" / "Commutative Ring"
   Restriction: Both A and B must be DEFINITIONS (not theorems). A must be a concrete/named object; B must be an abstract structure or class that A belongs to.

==== WHEN TO RETURN "none" ====

Return "none" if:
- The connection is only "both are about X" (e.g., both about primes, both about groups) without a structural relationship.
- The analogy is vague ("both establish invariants", "both are foundational").
- You would rate confidence below 0.75.

It is perfectly acceptable ~\textrm{\textemdash}~ and preferred ~\textrm{\textemdash}~ to return "none" often. A sparse, high-quality graph beats a dense noisy one.

==== FEW-SHOT EXAMPLES ====

Example 1 ~\textrm{\textemdash}~ strong generalizes:
  NODE A: "Binomial Theorem" (theorem, Combinatorics): (x+y)^n = ~$\Sigma$~ C(n,k) x^(n-k) y^k
  NODE B: "Sum of Binomial Coefficients" (theorem, Combinatorics): ~$\Sigma$~ C(n,k) = 2^n
  ~$\rightarrow$~ {"relationship": "generalizes", "confidence": 0.97, "reasoning": "Setting x=y=1 in the Binomial Theorem yields ~$\Sigma$~ C(n,k) = 2^n. B is literally a substitution instance of A."}

Example 2 ~\textrm{\textemdash}~ strong prerequisite:
  NODE A: "Prime Number" (definition, Number Theory): natural number p ~$\geq$~ 2 divisible only by 1 and p
  NODE B: "Fermat's Little Theorem" (theorem, Number Theory): if p prime, a^p ~$\equiv$~ a (mod p)
  ~$\rightarrow$~ {"relationship": "prerequisite", "confidence": 0.98, "reasoning": "The definition of prime number is required to state B: the hypothesis 'p prime' is essential and primality is used in the proof."}

Example 3 ~\textrm{\textemdash}~ "none" case (both about primes but no structural link):
  NODE A: "Wilson's Theorem" (theorem, Number Theory): (p-1)! ~$\equiv$~ -1 (mod p) for p prime
  NODE B: "Bertrand's Postulate" (theorem, Number Theory): for n ~$\geq$~ 1 there is a prime in (n, 2n]
  ~$\rightarrow$~ {"relationship": "none", "confidence": 1.0, "reasoning": "Both concern primes but share no structural relationship. Neither is used in the other's proof, neither generalizes the other, and they address unrelated questions (factorial modular identity vs. prime gap bound)."}

Example 4 ~\textrm{\textemdash}~ analogous-to (different domains, shared schema):
  NODE A: "Hensel's Lemma" (theorem, Number Theory): lifts a simple root of f mod p to a root in Z_p
  NODE B: "Implicit Function Theorem" (theorem, Analysis): lifts an approximate solution to an exact smooth solution under non-degeneracy
  ~$\rightarrow$~ {"relationship": "analogous-to", "confidence": 0.85, "reasoning": "Both are lifting theorems with the same schema: 'given an approximate solution and a non-degeneracy condition (nonzero derivative / Jacobian), a true solution exists and is unique locally'. Direct structural parallel across different domains."}

Example 5 ~\textrm{\textemdash}~ cross-domain-bridge:
  NODE A: "Vector Space" (definition, Linear Algebra): module over a field
  NODE B: "Field" (definition, Abstract Algebra): commutative ring where every nonzero element is invertible
  ~$\rightarrow$~ {"relationship": "cross-domain-bridge", "confidence": 0.92, "reasoning": "A vector space is defined as a module over a field; B's structure is the scalar domain of A. This is the foundational bridge between linear algebra and field theory."}

Example 6 ~\textrm{\textemdash}~ instance-of:
  NODE A: "Ring ZMod n" (definition, Number Theory): Z/nZ with addition and multiplication mod n
  NODE B: "Commutative Ring" (definition, Abstract Algebra): ring where multiplication is commutative
  ~$\rightarrow$~ {"relationship": "instance-of", "confidence": 0.95, "reasoning": "ZMod n is a concrete example of a commutative ring. The abstract structure B is instantiated by the concrete object A."}

Example 7 ~\textrm{\textemdash}~ applied-in:
  NODE A: "Fundamental Theorem of Arithmetic" (theorem, Number Theory): unique prime factorization
  NODE B: "Lifting the Exponent Lemma" (theorem, Number Theory): v_p(a^n - b^n) = v_p(a-b) + v_p(n) under conditions
  ~$\rightarrow$~ {"relationship": "applied-in", "confidence": 0.88, "reasoning": "LTE relies on p-adic valuations, which are well-defined only because FTA guarantees unique prime factorizations. FTA's result is invoked in B's proof."}

==== TASK ====

Classify the relationship from NODE A to NODE B (directional). Use Mathlib names/imports to assess structural proximity when useful.

Respond with EXACTLY this JSON (no other text):
{
  "relationship": "<one of: prerequisite | generalizes | specializes | equivalent-to | applied-in | analogous-to | cross-domain-bridge | instance-of | none>",
  "confidence": <0.0 to 1.0>,
  "reasoning": "<1-4 sentences. State the specific structural correspondence, not just 'both relate to X'.>"
}

If relationship is "none", set confidence to 1.0 (confident no relationship exists).
\end{promptbox}

\subsection{Edge Type Restrictions and Design Rationale}
\label{app:edge_restrictions}

Each of the eight edge types has node-type restrictions
(Table~\ref{tab:edge_restrictions}).  These restrictions are stated in the LLM
prompt (see the ontology block above) and serve two purposes: (1)~they reduce
misclassification by excluding structurally impossible combinations (e.g., specializes
is excluded between two definitions, because specialization is a
relationship between theorem statements); and (2)~they sharpen the distinction
between similar types, particularly prerequisite vs.\ applied-in.

\begin{table}[h]
\centering
\caption{Node-type restrictions for each edge type.  ``Defn'' = definition,
``Thm'' = theorem.  These constraints are stated verbatim in the LLM prompt and
enforced during post-processing.}
\label{tab:edge_restrictions}
\vspace{3pt}
\small
\begin{tabular}{llll}
\toprule
\textbf{Edge Type} & \textbf{$A$ must be} & \textbf{$B$ must be} & \textbf{Rationale} \\
\midrule
prerequisite        & defn    & any     & A concept needed to \emph{state} B \\
generalizes         & thm     & thm     & Strict containment between statements \\
specializes         & thm     & thm     & Inverse of generalizes \\
equivalent-to       & thm     & thm     & Mutual implication between statements \\
applied-in          & thm     & thm     & A's \emph{result} used in B's \emph{proof} \\
analogous-to        & thm     & thm     & Shared proof schema across domains or contexts \\
cross-domain-bridge & any     & any     & Domains must differ \\
instance-of         & defn    & defn    & Concrete object exemplifies abstract class \\
\bottomrule
\end{tabular}
\end{table}

\paragraph{Prerequisite vs.\ applied-in.}
These two types both capture ``$A$ supports $B$'' but differ in the nature of
the dependency:
\begin{itemize}
  \item prerequisite: $A$ is a \emph{definition} whose concept is
        required to \emph{formulate} $B$.  Without $A$, the statement of $B$
        cannot be written.  Example: the definition of ``prime number'' is a
        prerequisite for Fermat's Little Theorem, since the hypothesis ``$p$ is
        prime'' is part of the theorem statement.
  \item applied-in: $A$ is a \emph{theorem} whose result is invoked
        during the \emph{proof} of $B$.  The statement of $B$ can be written
        independently, but proving it requires $A$.  Example: Rolle's Theorem
        is applied in the standard proof of the Mean Value Theorem.
\end{itemize}
\noindent This separation ensures that definitional dependencies (which expand a
reader's vocabulary) are distinguished from proof-level dependencies (which
suggest proof strategies to the Prover agent).

\subsection{Post-Processing Edge Cleanup}
\label{app:edge_cleanup}

Although the node-type restrictions are stated verbatim in the prompt
(Table~\ref{tab:edge_restrictions}), 11.4\% of inferred edges (1{,}041 of 9{,}141)
violate them.  A programmatic post-processing pass enforces the
restrictions and removes the edges that violate them (Table~\ref{tab:edge_cleanup}).

\begin{table}[h]
\centering
\caption{Post-processing edge cleanup. ``Before'' counts are the raw LLM output before
restriction enforcement; ``Dropped'' counts are the edges violating the restrictions in
Table~\ref{tab:edge_restrictions}. The last two rows add the auto-inverse edges generated
in post-processing.}
\label{tab:edge_cleanup}
\vspace{3pt}
\small
\begin{tabular}{lrrrl}
\toprule
\textbf{Edge Type} & \textbf{Before} & \textbf{Dropped} & \textbf{After} & \textbf{Violation} \\
\midrule
applied-in          & 2{,}277 & 750 & 1{,}527 & $B$ (rarely $A$) is a definition \\
analogous-to        &    701 & 118 &    583 & at least one endpoint is a definition \\
cross-domain-bridge & 3{,}631 & 139 & 3{,}492 & same domain \\
generalizes         &     69 &  19 &     50 & $B$ is a definition \\
specializes         &     37 &   6 &     31 & $B$ is a definition \\
equivalent-to       &     17 &   4 &     13 & one endpoint is a definition \\
prerequisite        & 2{,}329 &   5 & 2{,}324 & $A$ is a theorem \\
instance-of         &     80 &   0 &     80 & no violations \\
\midrule
\textbf{Total (type violations)} & \textbf{9{,}141} & \textbf{1{,}041} & \textbf{8{,}100} & \\
\emph{Duplicate-ID edges (raw)} &  74 &  74 &      0 & edges related to two duplicate node IDs \\
\midrule
\textbf{Final raw total}    &        &        & \textbf{8{,}026} & \\
\midrule
\emph{Auto-inverse edges (generated)} & 1{,}417 & 9 & 1{,}408 & edges related to two duplicate node IDs \\
\textbf{Final graph}        &        &        & \textbf{9{,}434} & \\
\bottomrule
\end{tabular}
\end{table}

\noindent The largest source of violations is applied-in (750 edges, 72\% of all
violations), where an endpoint, usually $B$, is a definition. A definition has no proof, so
such an edge records no proof dependency; the reasoning texts typically describe the
definition's role in stating or proving the theorem (e.g., ``Pascal's Rule $\to$ Binomial
Coefficient''), which is the prerequisite relation in the opposite direction. The 139
cross-domain-bridge violations are same-domain pairs.

Of the 132{,}132 calls, 276 (0.2\%) failed after nine retries against a Bedrock daily
token quota and were treated as no relation (the affected pairs were not recorded).
An additional 74 raw edges were dropped because, at inference time, two node IDs were
each assigned to two theorems (\texttt{thm\_geom\_007} to Parallelogram Law and Thales'
Theorem; \texttt{thm\_geom\_008} to Orthocenter on Altitude and Riesz Representation), so
an edge on either ID could not be attributed to one theorem. The IDs were corrected in the
source data after inference. Removing these edges yields the final 8{,}026 raw edges
(per-type counts in \S\ref{sec:setup}).
The auto-inverse step generates 1{,}417 edges, of which 9 touch the duplicate IDs and
are dropped, leaving 1{,}408 and a final graph of \textbf{9{,}434 edges}.

Table~\ref{tab:kg_connectivity} reports connectivity in two ways: on average each node
has 51.8 directed edges but only 31.5 distinct neighbors (median 25), because the symmetric
types and the generalizes/specializes pair are stored as one edge in each direction, except
where the reverse direction already carried an edge of another type (351 edges: 319
cross-domain-bridge, 25 analogous-to, 7 equivalent-to), which the inverse step leaves as
returned. Of the
5{,}740 connected node pairs, 3{,}694 (64\%) are linked in both directions, from the model
returning both orderings or from auto-inversion; cross-domain-bridge, at 4{,}683 of the
9{,}434 edges, accounts for most of them. The 2{,}046 one-directional pairs are all
prerequisite, applied-in, or instance-of edges (1{,}400, 636, and 10), types without an
auto-inverse. Mutual pairs also occur: 390 carry applied-in in both directions and 288
carry prerequisite in both; post-processing removes contradictory pairs for generalizes and
specializes, and these edges remain as returned. Four nodes are isolated but remain
in the graph and the Explorer's candidate set.\footnote{All four stem from the
duplicate-ID fix: Parallelogram Law and Orthocenter on Altitude kept the shared IDs
and lost their ambiguous edges, while Thales' Theorem and Riesz Representation on Hilbert
spaces received new IDs after inference and so have no edges in this build.}

\begin{table}[h]
\centering
\caption{\MathKG{} connectivity statistics (final graph, 364 nodes, 9{,}434 directed
edges).}
\label{tab:kg_connectivity}
\vspace{3pt}
\small
\begin{tabular}{lr}
\toprule
\textbf{Statistic} & \textbf{Value} \\
\midrule
Directed edges per node (in $+$ out), mean & 51.8 \\
Distinct neighbors per node, mean          & 31.5 \\
Distinct neighbors per node, median        & 25 \\
Unique connected node pairs                & 5{,}740 \\
Isolated nodes                             & 4 \\
Highest-degree nodes (directed edges)      & Complex Numbers (254), Ring (253), Field (241) \\
\bottomrule
\end{tabular}
\end{table}

\section{Explorer and Prover Prompts}
\label{app:prompts}

\subsection{Explorer Prompts}
\label{app:explorer_prompts}

The Explorer makes two LLM calls per problem, both answered by the evaluated model
(Appendix~\ref{app:explorer}): seed scoring for Layer~1 (\texttt{with\_kg} and
\texttt{full\_system}) and search-term suggestion for Layer~2 (\texttt{with\_mathlib} and
\texttt{full\_system}). Each call sends the Explorer system prompt (first box) together with
its own user prompt (second and third boxes). Placeholders in braces are
filled per call; in the seed-scoring prompt, the candidate list holds one four-line entry per
candidate node.

\begin{promptbox}{Explorer system prompt}
You are an expert mathematician specializing in number theory and combinatorics. You excel at identifying structural patterns, analogies between mathematical domains, and gaps in mathematical knowledge. You reason carefully about formal mathematics.
\end{promptbox}

\begin{promptbox}{Seed-scoring prompt (Layer 1, one call per problem)}
Given this mathematical problem, score and select the {s} most relevant theorems from the candidates below.

PROBLEM:
{formal_statement}

DOMAIN: {domain}

CANDIDATE THEOREMS ({n} total):
- [{id}] {name}
  Statement: {statement}
  Keywords: {keywords}
  Mathlib: {lean4_mathlib_name}

Select exactly {s} theorems most relevant for solving this problem.
For each, explain WHY it's relevant (proof technique, key lemma, analogous structure, etc.).

Respond in JSON:
{
  "seeds": [
    {
      "theorem_id": "...",
      "name": "...",
      "relevance_score": 0.0-1.0,
      "rationale": "why this theorem is relevant"
    }
  ]
}
\end{promptbox}

\begin{promptbox}{Search-term suggestion prompt (Layer 2, one call per problem)}
Given this mathematical problem, suggest 10-15 Lean 4 Mathlib namespaces or identifier prefixes that would be useful to search for when proving it.

PROBLEM (natural language):
{informal_statement}

PROBLEM (formal Lean 4):
{formal_statement}

For each suggestion, classify its importance:
- "essential": directly needed for the proof
- "helpful": likely useful as a supporting lemma
- "speculative": might be relevant, worth checking

Focus on specific Mathlib namespaces and declaration families, e.g.:
- "Nat.Prime" (not just "prime")
- "Function.Injective" (not just "injective")
- "Finset.card" (not just "card")

Respond in JSON:
{
  "search_terms": [
    {"term": "Nat.Prime", "priority": "essential"},
    {"term": "Function.Injective", "priority": "helpful"},
    {"term": "Nat.strongRecOn", "priority": "speculative"}
  ]
}
\end{promptbox}

\subsection{Prover Prompts}
\label{app:prover_prompts}

The Prover uses three prompt components. On \textbf{attempt~$\boldsymbol{1}$}, the LLM receives the
system prompt plus the initial-generation prompt; on \textbf{attempts~$\boldsymbol{2}$--$\boldsymbol{K}$}, the system
prompt plus the refinement prompt, which carries the previous Lean code and up to ten Lean error
messages. When Lean reports unknown identifiers, a note listing them and forbidding their
reuse (\texttt{\{unknown\_id\_note\}} below) is appended to the errors. Placeholders in
braces are filled per call. The Explorer's context block (Appendix~\ref{app:explorer})
fills \texttt{\{proof\_sketch\}} and so appears only at attempt~1;
\texttt{\{context\}} is reserved for per-node formal statements, which the released
\MathKG{} does not include, so it carried no retrieval content in the reported runs.

\begin{promptbox}{Prover system prompt}
You are an expert in Lean 4 and Mathlib. You write correct, idiomatic Lean 4 proofs. You are familiar with Mathlib tactics: exact, apply, simp, ring, omega, norm_num, decide, have, obtain, rw, induction, rcases, constructor, push_neg, by_contra, linarith, nlinarith, polyrith, positivity, field_simp, gcongr, aesop, tauto, native_decide, and others.
IMPORTANT Lean 4 syntax rules:
1) Function application uses SPACES not parentheses: write `f x` not `f(x)`.
2) `import Mathlib` must be the VERY FIRST line ~\textrm{\textemdash}~ nothing before it.
3) Stop immediately once all goals are closed ~\textrm{\textemdash}~ do NOT add extra tactics after proof is done.
4) Use Lean 4 syntax ONLY, never Lean 3.
5) NEVER use `sorry` ~\textrm{\textemdash}~ it is not a valid proof and will be rejected. You must complete every proof with real tactics.
When given error messages, you diagnose them precisely and fix the proof.
\end{promptbox}

\begin{promptbox}{Initial-generation prompt (attempt 1)}
Write a complete Lean 4 proof for the following theorem.

NATURAL LANGUAGE STATEMENT:
{statement}

LEAN 4 TYPE SIGNATURE:
{lean4_sketch}

INFORMAL PROOF SKETCH:
{proof_sketch}

RELEVANT MATHLIB CONTEXT:
{context}

CRITICAL RULES:
1. The FIRST line MUST be `import Mathlib` ~\textrm{\textemdash}~ nothing before it.
2. Do NOT include any `import` statement other than the first line.
3. Use Lean 4 syntax ONLY. Function application uses spaces: `f x` not `f(x)`.
4. Stop immediately once all goals are solved ~\textrm{\textemdash}~ do NOT add extra tactics.
5. For inequality goals: try `linarith` first, then `nlinarith`, `polyrith`, or `positivity`.
6. For arithmetic: try `omega`, `norm_num`, or `ring`.
7. For decidable propositions on finite types: try `decide` or `native_decide`.
8. Only use Mathlib names you are CERTAIN exist. If a name does not exist, try a different proof strategy entirely.
9. Use Lean 4 syntax ONLY. Do NOT use Lean 3 syntax (no begin/end blocks). Use by followed by tactic lines.

Return ONLY the Lean 4 code, no explanation.
\end{promptbox}

\begin{promptbox}{Refinement prompt (attempts $\boldsymbol{2}$ to $\boldsymbol{K}$, after Lean returns errors)}
The following Lean 4 proof has errors. Fix it.

LEAN 4 CODE:
```lean
{previous_code}
```

ERRORS FROM LEAN:
{error_text}{unknown_id_note}

THEOREM TYPE SIGNATURE (must be preserved):
{lean4_sketch}

MATHLIB CONTEXT:
{context}

CRITICAL RULES:
- The FIRST line MUST be `import Mathlib` ~\textrm{\textemdash}~ nothing before it, no duplicate imports.
- Use Lean 4 syntax ONLY: function application uses spaces `f x` not `f(x)`.
- Stop immediately once all goals are solved ~\textrm{\textemdash}~ do NOT add extra tactics.

Common fixes:
- If 'unknown identifier' / 'unknown constant': the name does not exist in Mathlib.
  Try a different lemma or a different proof strategy entirely.
- If 'type mismatch': add explicit type annotations or use `show` tactic.
- If 'No goals to be solved' / 'goals accomplished': you have EXTRA tactics after
  the proof is done. REMOVE all tactics after the point where goals are closed.
- If 'unexpected token' / 'unexpected identifier': you have a SYNTAX error.
  Check: no Lean 3 syntax, correct indentation, correct keyword usage.
- If 'Function expected': you are applying something that is not a function.
  Check parentheses and application syntax.
- If 'failed to synthesize': add missing typeclass instances with `@[instance]` or `haveI`.
- If 'linarith failed': try `nlinarith`, `polyrith`, or `positivity` instead.
- For arithmetic goals: try `omega`, `norm_num`, `ring`, or `field_simp`.
- For decidable propositions: try `decide` or `native_decide`.
- Only use Mathlib names you are CERTAIN exist. If a name doesn't exist, try a different lemma or a different proof strategy entirely.
9. Use Lean 4 syntax ONLY. Do NOT use Lean 3 syntax (no begin/end blocks). Use by followed by tactic lines.

Return ONLY the corrected Lean 4 code, no explanation.
\end{promptbox}

\section{Implementation and Compute Setup}
\label{app:implementation}

\paragraph{Agent Architecture.}
The \MathAgent{} system is implemented from scratch without external agentic
frameworks (e.g., AutoGen, LangChain, CrewAI), using standard Python clients for the
three LLM backends and NetworkX for the graph. All agents share one LLM interface that
sends a system prompt and a user prompt to Claude via AWS Bedrock or the Anthropic API, or
to a local vLLM endpoint for the open-weight models, and a JSON parser that accepts a JSON
object whether it is the entire reply, inside a Markdown code fence, or embedded in
surrounding text.

\paragraph{Communication Pattern.}
Agents communicate through:
\begin{itemize}
  \item \textbf{Shared graph}: all agents work on the same \MathKG{} graph object (NetworkX).
  \item \textbf{Structured output}: each agent returns a dictionary with a fixed set of
        fields.
  \item \textbf{Sequential execution}: for each problem the Explorer builds the context
        and the Prover then runs its attempt loop; parallelism is across problems only.
\end{itemize}

\paragraph{API Configuration.}
Claude Sonnet 4.6 via an AWS Bedrock inference profile.
Temperature: 0.3 for initial proof generation, 0.2 for refinement attempts. Thinking mode is
disabled for the Qwen3 and Goedel-Prover-V2 models.
Maximum output tokens are configured per model:
Claude~Sonnet~4.6, Qwen3-8B, and Qwen3-32B use 2{,}000; Goedel-Prover-V2-8B uses 16{,}384;
and Goedel-Prover-V2-32B uses 4{,}096. Each cap exceeds the length of that model's
longest verified proof. LLM calls are retried with exponential backoff (5 retries, 10--160\,s
delays) to recover from transient API failures.

\paragraph{Graph Representation.}
\MathKG{} is held in memory as a NetworkX \texttt{DiGraph} with one node per theorem or
definition. Each node carries six attributes
(\texttt{id}, \texttt{name}, \texttt{domain}, \texttt{statement}, \texttt{type},
\texttt{keywords}); each edge carries its type, confidence, and reasoning, and generated
inverse edges are marked as such.
The released graph file (364 nodes, 9{,}434 edges, 7.0\,MB) is plain JSON with a
\texttt{nodes} list and an \texttt{edges} list; it is the graph the \texttt{with\_kg} and
\texttt{full\_system} runs searched. The
node source file additionally stores each node's \Mathlib{} declaration name, import, and
verification status; the released graph carries the six attributes above
(Appendix~\ref{app:mathkg_node}).

\paragraph{Compute Infrastructure.}
\emph{Sonnet~4.6 ablation:} AWS EC2 \texttt{r7i.48xlarge}
(192~vCPU, 1{,}536\,GB RAM), 20 parallel workers.
Lean~4 v4.30.0-rc2 with \Mathlib{}~v4.30.0-rc2. Each Lean compilation
is allowed up to 1{,}200\,s (20~minutes) of wall-clock time and 200\,GB of virtual memory;
a compilation exceeding either limit is terminated and counted as a failed attempt.

\emph{Qwen3-8B ablation:} AWS EC2 \texttt{g6e.24xlarge}
(96~vCPU, 768\,GB RAM, 4$\times$ NVIDIA L40S GPUs), 15 parallel workers.
Model served via vLLM in BF16, the precision of the released weights. Same Lean~4
timeout and memory limit as in all runs (1{,}200\,s, 200\,GB).

\emph{Goedel-Prover-V2 (32B, 8B) and Qwen3-32B ablations:} AWS EC2
\texttt{g7e.48xlarge} (192~vCPU, 2{,}048\,GB RAM, 8$\times$ NVIDIA RTX~PRO~6000
Blackwell, 96\,GB each), with 64 parallel workers for the
Goedel-Prover-V2 runs and 96 for Qwen3-32B. All three models were served via vLLM~0.25.0 (PyTorch~2.11, CUDA~13) in BF16, the
precision of the released weights, with one full-model replica per GPU and the models'
native 40{,}960-token context window. Verification uses Lean~4 v4.32.0-rc1 with
\Mathlib{}~v4.32.0-rc1 under the same Lean timeout and memory limit (1{,}200\,s, 200\,GB).

\paragraph{Code Extraction and Verification.}
The same rule applies to every model: the last fenced \texttt{lean4} code block in a
reply is taken as the proof; otherwise the reply is cut from its first \texttt{import},
\texttt{theorem}, or \texttt{def} line onward. An attempt then counts as
solved only if all three checks pass: (1)~\emph{no sorry}: a Lean warning containing
``sorry'' fails the attempt; (2)~\emph{statement fidelity}: the file must contain the
assigned theorem with its statement, and every declaration the benchmark supplies,
identical to the benchmark text up to whitespace; a file that compiles but alters, omits,
or replaces them is not a solve; (3)~\emph{clean compilation}: a non-zero exit code or
any error diagnostic fails the attempt.

Some PutnamBench problems ask for a closed-form answer in addition to a proof. Their Lean
files contain an answer slot (an \texttt{abbrev} ending in \texttt{:= sorry}), and the
benchmark supplies the correct answer in an adjacent comment. Following the benchmark's
standard setting, the loader writes that answer into the slot before the model sees the
problem, so the model proves the theorem with the answer fixed, and check~(2) rejects any
proof that alters it. Results for the 346 answer-slot problems come from a dedicated run with
the same model, prompts, attempt budget, and decoding parameters as the Sonnet~4.6 ablation of
Appendix~\ref{app:sonnet_ablation}; results for the other 326 problems, which have no slot,
come from that ablation itself.

\paragraph{Code Availability.}
The complete implementation (agents, Orchestrator, graph construction, benchmark
evaluation), the \MathKG{} graph and node files, and the per-problem results of all 28
evaluation runs
\ificlrfinal
are released at \href{https://github.com/sarehnabi/mathagent}{\textcolor{blue}{\nolinkurl{https://github.com/sarehnabi/mathagent}}} under the MIT license.
\else
will be released under the MIT license upon publication.
\fi

\section{Example Generated Proofs}
\label{app:proofs}

We show two verified proofs of the same miniF2F problem,
\texttt{mathd\_numbertheory\_335} (validation split): if $n \bmod 7 = 5$, then
$5n \bmod 7 = 4$, produced in \texttt{prover\_only} on the first attempt by
Claude Sonnet~4.6 and by Goedel-Prover-V2-8B.

\begin{codebox}{Claude Sonnet 4.6, \texttt{prover\_only}, attempt 1}
import Mathlib

theorem mathd_numbertheory_335
  (n : @$\mathbb{N}$@)
  (h@$_{0}$@ : n % 7 = 5) :
  (5 * n) % 7 = 4 := by
  omega
\end{codebox}

\noindent Sonnet closes the goal with a single decision procedure. Goedel-Prover-V2-8B,
below, reaches the same goal through nested \texttt{have} steps, with the arithmetic done by one
\texttt{simp} rewrite followed by \texttt{norm\_num}. Both proofs pass the checks of
Appendix~\ref{app:implementation}, and both models solved this problem in all four augmentation modes.

\begin{codebox}{Goedel-Prover-V2-8B, \texttt{prover\_only}, attempt 1}
import Mathlib

theorem mathd_numbertheory_335
  (n : @$\mathbb{N}$@)
  (h@$_{0}$@ : n % 7 = 5) :
  (5 * n) % 7 = 4 := by
  have h@$_{1}$@ : (5 * n) % 7 = 4 := by
    have h@$_{2}$@ : n % 7 = 5 := h@$_{0}$@
    have h@$_{3}$@ : (5 * n) % 7 = 4 := by
      have h@$_{4}$@ : n % 7 = 5 := h@$_{2}$@
      have h@$_{5}$@ : (5 * n) % 7 = 4 := by
        -- Use the fact that n @$\equiv$@ 5 mod 7 to simplify the expression
        have h@$_{6}$@ : n % 7 = 5 := h@$_{4}$@
        have h@$_{7}$@ : (5 * n) % 7 = 4 := by
          -- Use the property of modular arithmetic
          -- to simplify the expression
          have h@$_{8}$@ : n % 7 = 5 := h@$_{6}$@
          have h@$_{9}$@ : (5 * n) % 7 = (5 * (n % 7)) % 7 := by
            simp [Nat.mul_mod, Nat.add_mod, Nat.mod_mod]
          rw [h@$_{9}$@, h@$_{8}$@]
          <;> norm_num
        exact h@$_{7}$@
      exact h@$_{5}$@
    exact h@$_{3}$@
  exact h@$_{1}$@
\end{codebox}

\section{Sonnet Ablation Across Benchmarks}
\label{app:sonnet_ablation}

This appendix gives the full three-benchmark ablation behind \S\ref{sec:complementarity},
using Sonnet~4.6. Table~\ref{tab:sonnet_bench} reports the four-mode results with the
per-problem oracle (the union of problems solved by any of the four modes):
each augmented mode lowers the three-benchmark aggregate by 2.1--3.9\% relative to
\texttt{prover\_only}, while the oracle exceeds \texttt{prover\_only} by 6\% to 32\%, most on
PutnamBench.

\begin{table}[h]
\centering
\caption{Sonnet~4.6 full four-mode ablation across three benchmarks (pass@32), with the
per-problem \emph{oracle} (union of all four modes). Best fixed mode per row in \textbf{bold};
the oracle exceeds \texttt{prover\_only} by $+6\%$ to $+32\%$, most on PutnamBench.}
\label{tab:sonnet_bench}
\vspace{3pt}
\small
\setlength{\tabcolsep}{4pt}
\resizebox{\textwidth}{!}{%
\begin{tabular}{lcccccc}
\toprule
\textbf{Benchmark} & \textbf{prover\_only} & \textbf{with\_kg} & \textbf{with\_mathlib} & \textbf{full\_system} & \textbf{oracle} & \textbf{$\Delta$ vs.\ prover\_only} \\
\midrule
MathOlympiadBench (360) & \textbf{39 (10.8\%)} & 33 (9.2\%) & 36 (10.0\%) & 38 (10.6\%) & 50 (13.9\%) & $+11$ ($+28\%$) \\
miniF2F (488)           & 365 (74.8\%) & 358 (73.4\%) & 359 (73.6\%) & \textbf{366 (75.0\%)} & 388 (79.5\%) & $+23$ ($+6\%$) \\
PutnamBench (672)       & \textbf{34 (5.1\%)} & 30 (4.5\%) & 31 (4.6\%) & 25 (3.7\%) & 45 (6.7\%) & $+11$ ($+32\%$) \\
\midrule
\textbf{Total (1{,}520)} & \textbf{438 (28.8\%)} & 421 (27.7\%) & 426 (28.0\%) & 429 (28.2\%) & \textbf{483 (31.8\%)} & $+45$ ($+10\%$) \\
\bottomrule
\end{tabular}%
}
\end{table}

Table~\ref{tab:sonnet_unique} reports, per
benchmark, how many problems each augmented mode solves \emph{beyond} \texttt{prover\_only},
together with the oracle and the problems lost (solved by
\texttt{prover\_only} but by no augmented mode); Figure~\ref{fig:venn_main} shows the
disjoint overlap structure.

\begin{table}[h]
\centering
\caption{Sonnet~4.6 problem-level complementarity. \texttt{+KG}/\texttt{+Mathlib}/%
\texttt{+Full}: problems each mode solves beyond \texttt{prover\_only} (the modes overlap;
disjoint regions are in Figure~\ref{fig:venn_main}). Oracle: solved by any of the four
modes. Lost: solved only by \texttt{prover\_only}.}
\label{tab:sonnet_unique}
\vspace{3pt}
\small
\begin{tabular}{lccccc}
\toprule
\textbf{Benchmark} & \textbf{prover\_only} & \textbf{+KG} & \textbf{+Mathlib} & \textbf{+Full} & \textbf{Oracle / Lost} \\
\midrule
MathOlympiadBench & 39 & +6 & +4 & +9 & 50 / $-3$ \\
miniF2F & 365 & +10 & +9 & +18 & 388 / $-7$ \\
PutnamBench & 34 & +5 & +4 & +4 & 45 / $-3$ \\
\bottomrule
\end{tabular}
\end{table}

Aggregated across the three benchmarks, 45 problems are solved by at least one augmented
mode but not by \texttt{prover\_only}, versus 13 solved only by \texttt{prover\_only} and
lost by every augmented mode, roughly 3:1 in favor of augmentation at the problem
level, even though aggregate counts favor the baseline. The oracle exceeds
\texttt{prover\_only} by 6\% on miniF2F, 28\% on MathOlympiadBench, and 32\% on
PutnamBench: complementarity strengthens on harder problems.

\paragraph{Pass@$k$ analysis.}
Table~\ref{tab:passk} breaks down the Sonnet ablation by attempt budget. On miniF2F,
\texttt{with\_mathlib} leads at pass@1 ($+12$ over \texttt{prover\_only}), but
\texttt{prover\_only} leads from pass@3 to pass@20, with the augmented modes trailing by 9
to 15 problems at pass@3 and pass@5; at pass@32 \texttt{with\_kg} and
\texttt{with\_mathlib} trail by 7 and 6 while \texttt{full\_system} edges ahead by one. On PutnamBench,
\texttt{with\_kg} leads at pass@1 and pass@10 by 1 and 3, \texttt{prover\_only} leads
elsewhere, and at pass@32 the augmented modes trail by 3 to 9. On MathOlympiadBench the counts are
small and the modes trade places (\texttt{with\_mathlib} leads at pass@5,
\texttt{full\_system} at pass@10 and pass@20), with \texttt{prover\_only} ahead by 1 to 6 at pass@32.

\begin{table}[h]
\centering
\caption{Sonnet~4.6 pass@$k$ solve counts (solve rate); best mode per row in \textbf{bold}.}
\label{tab:passk}
\vspace{3pt}
\begin{tabular}{lcccc}
\toprule
\textbf{pass@$k$} & \textbf{prover\_only} & \textbf{with\_kg} & \textbf{with\_mathlib} & \textbf{full\_system} \\
\midrule
\multicolumn{5}{l}{\emph{miniF2F (488 problems)}} \\[2pt]
pass@1  & 198 (40.6\%) & 201 (41.2\%) & \textbf{210 (43.0\%)} & 206 (42.2\%) \\
pass@3  & \textbf{299 (61.3\%)} & 284 (58.2\%) & 290 (59.4\%) & 290 (59.4\%) \\
pass@5  & \textbf{326 (66.8\%)} & 311 (63.7\%) & 311 (63.7\%) & 316 (64.8\%) \\
pass@10 & \textbf{349 (71.5\%)} & 339 (69.5\%) & 339 (69.5\%) & 341 (69.9\%) \\
pass@20 & \textbf{358 (73.4\%)} & 352 (72.1\%) & 352 (72.1\%) & 352 (72.1\%) \\
pass@32 & 365 (74.8\%) & 358 (73.4\%) & 359 (73.6\%) & \textbf{366 (75.0\%)} \\[3pt]
\midrule
\noalign{\vskip 3pt}\multicolumn{5}{l}{\emph{PutnamBench (672 problems)}} \\[2pt]
pass@1  & 6 (0.9\%) & \textbf{7 (1.0\%)} & 6 (0.9\%) & 5 (0.7\%) \\
pass@3  & \textbf{15 (2.2\%)} & 13 (1.9\%) & 9 (1.3\%) & 11 (1.6\%) \\
pass@5  & \textbf{17 (2.5\%)} & 16 (2.4\%) & 11 (1.6\%) & 14 (2.1\%) \\
pass@10 & 21 (3.1\%) & \textbf{24 (3.6\%)} & 18 (2.7\%) & 21 (3.1\%) \\
pass@20 & \textbf{28 (4.2\%)} & 25 (3.7\%) & 24 (3.6\%) & 25 (3.7\%) \\
pass@32 & \textbf{34 (5.1\%)} & 30 (4.5\%) & 31 (4.6\%) & 25 (3.7\%) \\[3pt]
\midrule
\noalign{\vskip 3pt}\multicolumn{5}{l}{\emph{MathOlympiadBench (360 problems)}} \\[2pt]
pass@1  & \textbf{6 (1.7\%)} & 3 (0.8\%) & \textbf{6 (1.7\%)} & 4 (1.1\%) \\
pass@3  & \textbf{11 (3.1\%)} & 5 (1.4\%) & \textbf{11 (3.1\%)} & 8 (2.2\%) \\
pass@5  & 16 (4.4\%) & 13 (3.6\%) & \textbf{18 (5.0\%)} & 14 (3.9\%) \\
pass@10 & 25 (6.9\%) & 22 (6.1\%) & 25 (6.9\%) & \textbf{26 (7.2\%)} \\
pass@20 & 33 (9.2\%) & 28 (7.8\%) & 33 (9.2\%) & \textbf{34 (9.4\%)} \\
pass@32 & \textbf{39 (10.8\%)} & 33 (9.2\%) & 36 (10.0\%) & 38 (10.6\%) \\
\bottomrule
\end{tabular}
\end{table}

\section{Context Overflow (Goedel-Prover-V2-32B)}
\label{app:overflow}

Context overflow appeared only in the Goedel-Prover-V2-32B runs, and only with KG
augmentation. The KG context block is uncapped (Appendix~\ref{app:explorer}),
and for some problems the KG-augmented prompt plus the 4{,}096-token generation budget
exceeded the served 40{,}960-token context window, so the vLLM server rejected the request.
The client then lowers the generation budget to the room the prompt leaves and resends the
same prompt, so the model still sees the full augmentation context. When the
prompt alone filled the window, every retry failed, and after five retries the evaluator
recorded the problem as an error, counted as a non-solve. This affected only the two KG
modes: 10 problems under \texttt{with\_kg} and 28 under \texttt{full\_system} (one problem
in both), while \texttt{prover\_only} and \texttt{with\_mathlib} saw none. No other model's
runs recorded an overflow.

In Table~\ref{tab:xmodel_main} the overflowed problems are counted as failures against the
full denominator of 488, giving Goedel-32B 234/488 (48.0\%) for \texttt{with\_kg} and
208/488 (42.6\%) for \texttt{full\_system}. Excluding them, the rates are 234/478 (49.0\%)
and 208/460 (45.2\%), while \texttt{prover\_only} and \texttt{with\_mathlib} are unchanged
at 237/488 (48.6\%) and 235/488 (48.2\%). Restricting \texttt{prover\_only} to the same
problems gives 233/478 and 214/460, so the \texttt{with\_kg} gap of
Table~\ref{tab:xmodel_decomp} moves from $-3$ to $+1$, and the \texttt{full\_system} gap
closes from $-29$ to $-6$ against \texttt{prover\_only} and, in Table~\ref{tab:xmodel_decomp}'s
KG-beyond-Mathlib row, from $-27$ to $-3$ against \texttt{with\_mathlib}.

\section{Significance Tests for Table~\ref{tab:xmodel_decomp}}
\label{app:mcnemar}

Table~\ref{tab:mcnemar} reports, for every delta in Table~\ref{tab:xmodel_decomp}, the
$p$-value of a two-sided exact McNemar test on the discordant problems, those solved under
one mode and not the other. Six deltas are significant at $p<0.05$; every Sonnet delta and
every \Mathlib{} $-$ \texttt{prover\_only} delta is within noise.

\begin{table}[h]
\centering
\caption{Solve-count deltas of Table~\ref{tab:xmodel_decomp} with McNemar $p$-values (miniF2F, 488 problems).}
\label{tab:mcnemar}
\vspace{3pt}
\small
\setlength{\tabcolsep}{4.5pt}
\begin{tabular}{lccccc}
\toprule
\textbf{Question} & \textbf{Qwen3-8B} & \textbf{Qwen3-32B} & \textbf{Goedel-8B} & \textbf{Goedel-32B} & \textbf{Sonnet} \\
\midrule
KG helps? (kg $-$ PO)                 & $+2$ (.79) & $-14$ (.007) & $+14$ (.029) & $-3$ (.76) & $-7$ (.25) \\
Mathlib helps? (mathlib $-$ PO)       & $-2$ (.82) & $0$ (1.0)    & $-1$ (1.0)   & $-2$ (.86) & $-6$ (.31) \\
KG beyond Mathlib? (full $-$ mathlib) & $+5$ (.30) & $-16$ (.002) & $+1$ (1.0)   & $-27$ ($<$.001) & $+7$ (.28) \\
Synergy? (full $-$ max(kg,mathlib))   & $+1$ (1.0) & $-16$ (.002) & $-14$ (.024) & $-27$ ($<$.001) & $+7$ (.28) \\
\bottomrule
\end{tabular}
\end{table}

\section{Oracle at a Matched Attempt Budget}
\label{app:matched_oracle}

Table~\ref{tab:matched_oracle} compares the oracle of Table~\ref{tab:xmodel_oracle} with a
matched-budget variant that gives each of the four augmentation modes 8 attempts, 32 in total, the same
number \texttt{prover\_only} receives at pass@32; a problem counts for the matched-budget
union if any mode solves it within its first 8 attempts.

\begin{table}[h]
\centering
\caption{Oracle at a matched attempt budget on miniF2F (488 problems). Cells give solve counts
with the solve rate in parentheses; the header gives total attempts in parentheses. Matched
oracle: 8 attempts per mode ($4{\times}8{=}32$). Full oracle: 32 attempts per mode
($4{\times}32{=}128$), as in Table~\ref{tab:xmodel_oracle}. Gains: oracle minus
\texttt{prover\_only}, in problems (solve-rate points).}
\label{tab:matched_oracle}
\vspace{3pt}
\small
\setlength{\tabcolsep}{3pt}
\begin{tabular}{lccccc}
\toprule
\textbf{Model} & \textbf{prover\_only (32)} & \textbf{Matched oracle (32)} & \textbf{Matched gain} & \textbf{Full oracle (128)} & \textbf{Full gain} \\
\midrule
Qwen3-8B   & 33 (6.8\%)   & 51 (10.5\%)  & $+18$ ($+3.7$) & 52 (10.7\%)  & $+19$ ($+3.9$) \\
Qwen3-32B  & 62 (12.7\%)  & 79 (16.2\%)  & $+17$ ($+3.5$) & 80 (16.4\%)  & $+18$ ($+3.7$) \\
Goedel-8B  & 205 (42.0\%) & 232 (47.5\%) & $+27$ ($+5.5$) & 241 (49.4\%) & $+36$ ($+7.4$) \\
Goedel-32B & 237 (48.6\%) & 259 (53.1\%) & $+22$ ($+4.5$) & 266 (54.5\%) & $+29$ ($+5.9$) \\
Sonnet~4.6 & 365 (74.8\%) & 370 (75.8\%) & $+5$ ($+1.0$)  & 388 (79.5\%) & $+23$ ($+4.7$) \\
\bottomrule
\end{tabular}
\end{table}

\end{document}